\documentclass{article}

\PassOptionsToPackage{round}{natbib}

\usepackage[preprint]{neurips_2025}

\usepackage[utf8]{inputenc} %
\usepackage[T1]{fontenc}    %
\usepackage{hyperref}       %
\usepackage{url}            %
\usepackage{booktabs}       %
\usepackage{amsfonts}       %
\usepackage{nicefrac}       %
\usepackage{microtype}      %
\usepackage[table,xcdraw]{xcolor}         %
\usepackage{amsmath}
\usepackage{wrapfig}
\usepackage{graphicx}
\usepackage{subcaption}
\usepackage{amsthm}
\usepackage{xspace}
\usepackage{amssymb}
\usepackage{pifont}
\usepackage[colorinlistoftodos, textsize=small]{todonotes}
\usepackage[normalem]{ulem}
\useunder{\uline}{\ul}{}
\usepackage{multirow}
\usepackage{siunitx}
\usepackage{enumitem}
\usepackage{tablefootnote}
\usepackage{threeparttable}

\usepackage{tcolorbox}
\usepackage{fancyvrb}
\usepackage{fvextra}

\usepackage{siunitx}
\newcommand{\ulsi}[1]{#1\llap{\rule[-0.8ex]{\widthof{#1}}{0.5pt}}}

\newcommand{\agentbench}{\textsc{AgentBench}\xspace}
\newcommand{\benchmark}{\textsc{AgentBench-fc}\xspace}
\newcommand{\model}{\textsc{AgentRL}\xspace}
\newcommand{\xmark}{\ding{55}}
\newcommand{\err}[1]{\textcolor{gray}{$_{\pm#1}$}}
\newcommand{\vpara}[1]{\vspace{0.04in}\noindent\textbf{#1}\xspace}
\definecolor{mydarkgreen}{RGB}{0,80,0}
\newcommand{\cm}{\textcolor{mydarkgreen}{\checkmark}}
\newcommand{\xm}{\textcolor{black}{\xmark}}

\definecolor{darkred}{HTML}{8B0000}

\newcommand{\hide}[1] %

\title{\model: Reinforcing Multi-task LLM Agents\\From Zero}
\title{\model: Scaling Agentic Reinforcement Learning with a Multi-Turn, Multi-Task Framework}

\author{Hanchen Zhang$^{1\dagger*}$, Xiao Liu$^{1,2*}$, Bowen Lv$^{1\dagger}$, Xueqiao Sun$^{1}$, Bohao Jing$^{2\dagger}$, \\
\textbf{Iat Long Iong$^{1\dagger}$, Zhenyu Hou$^{1\dagger}$, Zehan Qi$^{1\dagger}$, Hanyu Lai$^{1\dagger}$, Yifan Xu$^{1\dagger}$, Rui Lu$^{1\dagger}$, } \\
\textbf{Hongning Wang$^{1}$, Jie Tang$^{1}$, Yuxiao Dong$^{1}$} \\
\\
\textsuperscript{1} Tsinghua University \quad
\textsuperscript{2} Z.AI \quad
}

\begin{document}

\maketitle

\renewcommand{\thefootnote}{\fnsymbol{footnote}}
    \footnotetext[1]{HZ and XL contributed equally.}
    \footnotetext[2]{Work done while these authors interned at Z.AI.}
\renewcommand{\thefootnote}{\arabic{footnote}}

\begin{abstract}
Recent advances in large language models (LLMs) have sparked growing interest in building generalist agents that can learn through online interactions. 
However, applying reinforcement learning (RL) to train LLM agents in multi-turn, multi-task settings remains challenging due to lack of  scalable infrastructure and stable training algorithms. 
In this work, we present the \model framework for scalable multi-turn, multi-task agentic RL training. 
On the infrastructure side, \model features a fully-asynchronous generation-training pipeline for efficient multi-turn RL. 
To support heterogeneous environment development in multi-task RL, we design a unified function-call based API interface, containerized environment development, and a centralized controller.   
On the algorithm side, we propose cross-policy sampling to encourage model exploration in multi-turn settings and task advantage normalization to stabilize multi-task training. 
Experiments show that \model, trained on open LLMs across five agentic tasks, significantly  outperforms GPT-5, Clause-Sonnet-4,  DeepSeek-R1, and other open-source LLM agents. 
Multi-task training with \model matches the best results among all task-specific models. 
\model is open-sourced at 
\url{https://github.com/THUDM/AgentRL}.
The algorithm and framework are adopted in building \textsc{\href{https://autoglm.zhipuai.cn}{AutoGLM}}~\citep{liu2024autoglmautonomousfoundationagents}.
\end{abstract}

\vspace{-3mm}
\begin{figure}[h]
  \centering
  \begin{subfigure}[c]{0.4\linewidth}
    \centering
    \includegraphics[width=\linewidth]{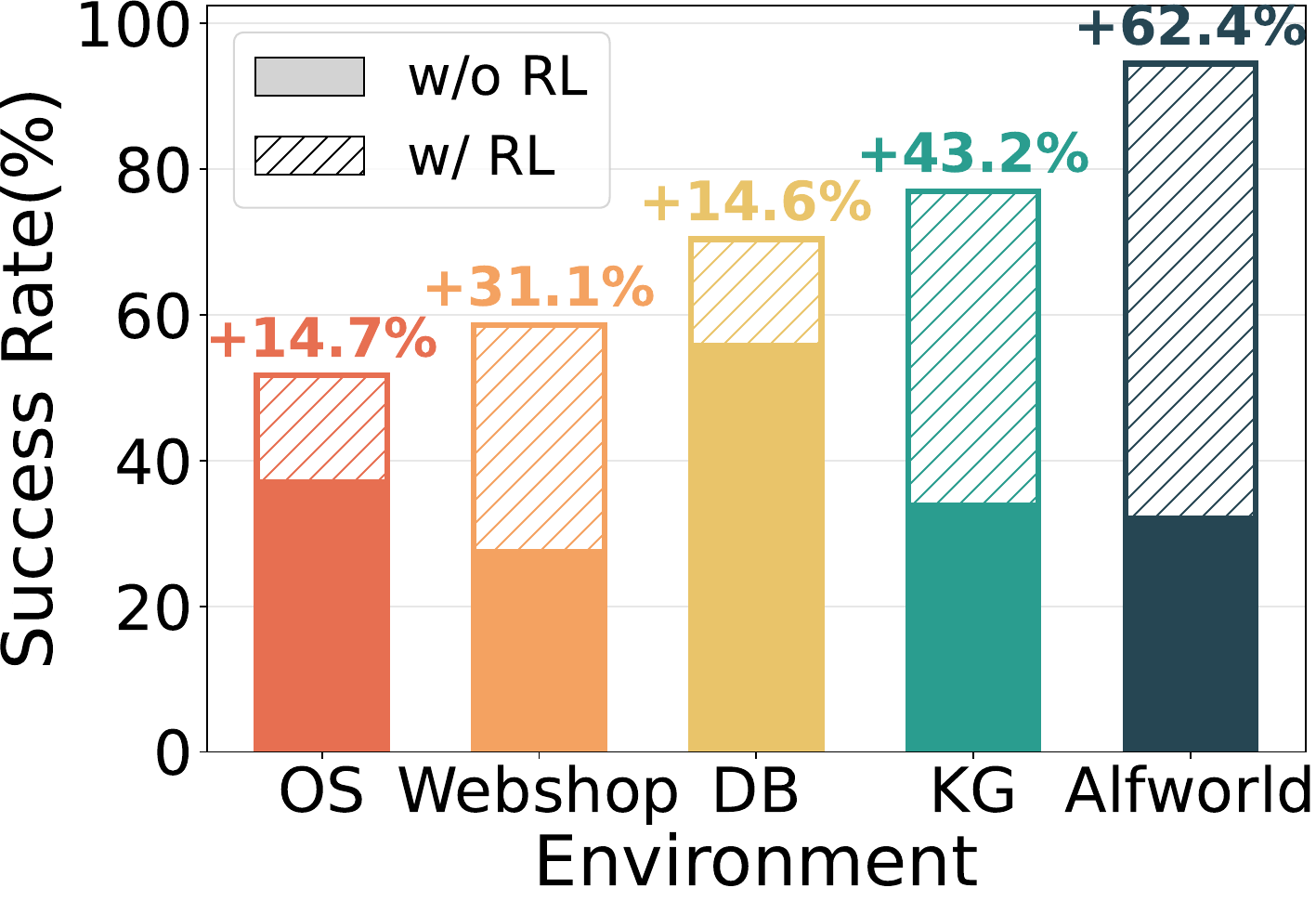}
    \caption{Gains of \model over the base (32B).}
  \end{subfigure}
  \quad
  \begin{subfigure}[c]{0.4\linewidth}
    \centering
    \includegraphics[width=\linewidth]{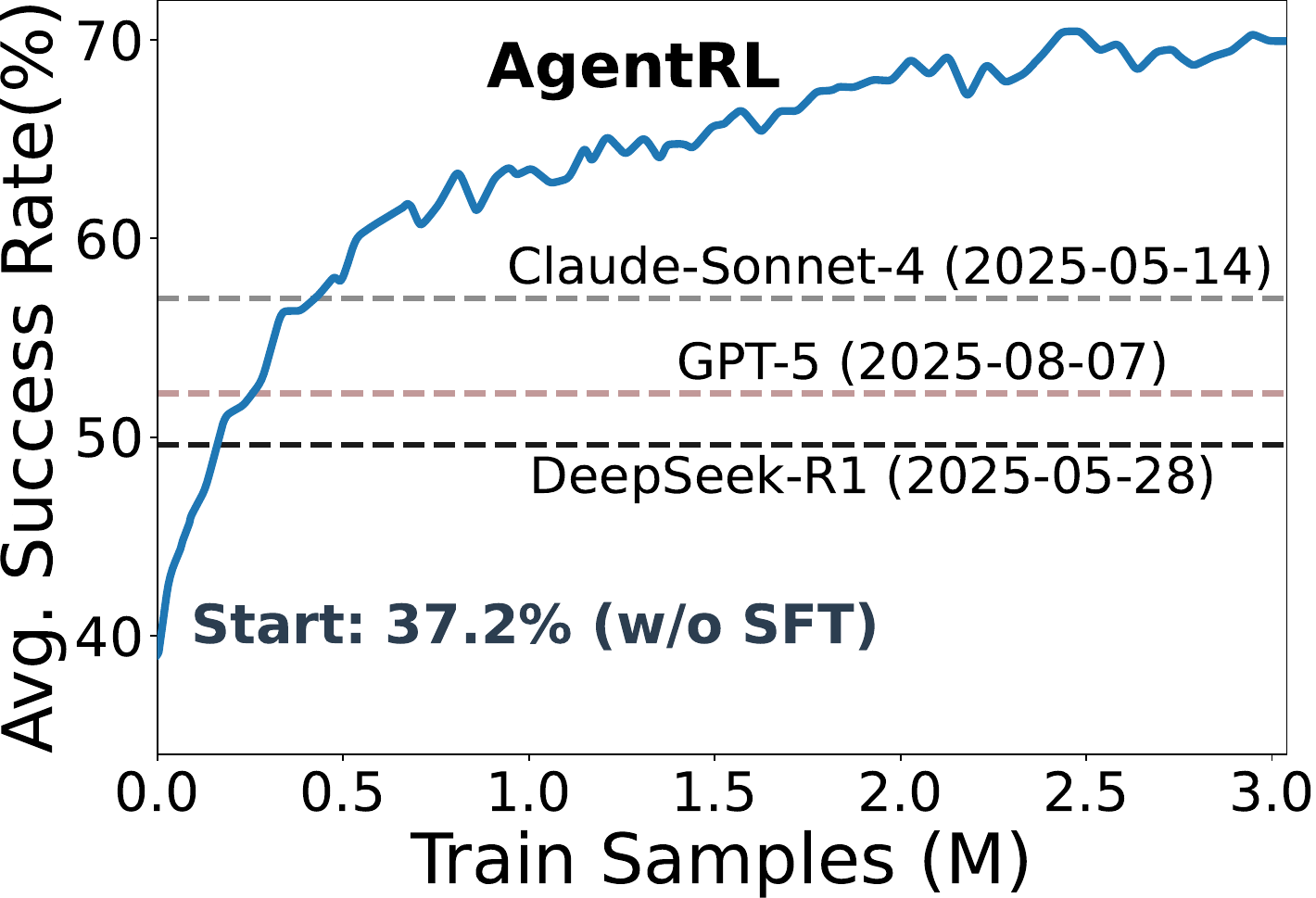}
    \caption{RL progress of \model (32B).}
    \label{fig:curve}
  \end{subfigure}
\vspace{-2mm}
  \caption{Overall performance of \model.}
  \label{fig:first}
\end{figure}
\vspace{-5mm}

\hide{

\begin{abstract}
Recent advances in large language models (LLMs) have sparked growing interest in building general-purpose agents that can learn through online interactions. However, applying reinforcement learning (RL) to train LLM agents in multi-turn, multi-task settings remains challenging due to limited exploration capacity, generalization degradation, and a lack of scalable infrastructure. In this work, we present a new framework for scalable online RL with LLM agents. Our method introduces a cross-policy sampling strategy  that improves exploration by sampling actions from a pool of models to enhance sample diversity and learning efficiency, and a task advantage normalization method that applies normalization over the advantage of task-level traces. We pair this with a fully asynchronous, decoupled training architecture that maximizes hardware throughput while preserving training stability. To support this system, we design a new agent environment backend, offering a unified API, high-performance controller, and full support for containerized rollout execution. Experiments on multi-task LLM agent benchmarks demonstrate that our approach achieves state-of-the-art performance, with improved sample efficiency, robustness, and generalization across diverse real-world tasks. 
The algorithm and framework are adopted in building \textsc{\href{https://autoglm.zhipuai.cn}{AutoGLM}}~\citep{liu2024autoglmautonomousfoundationagents}.
The framework is open-sourced at 
\url{https://github.com/THUDM/AgentRL}.
\end{abstract}

\vspace{-5mm}
\begin{figure}[h]
  \centering
  \begin{subfigure}[c]{0.45\linewidth}
    \centering
    \includegraphics[width=\linewidth]{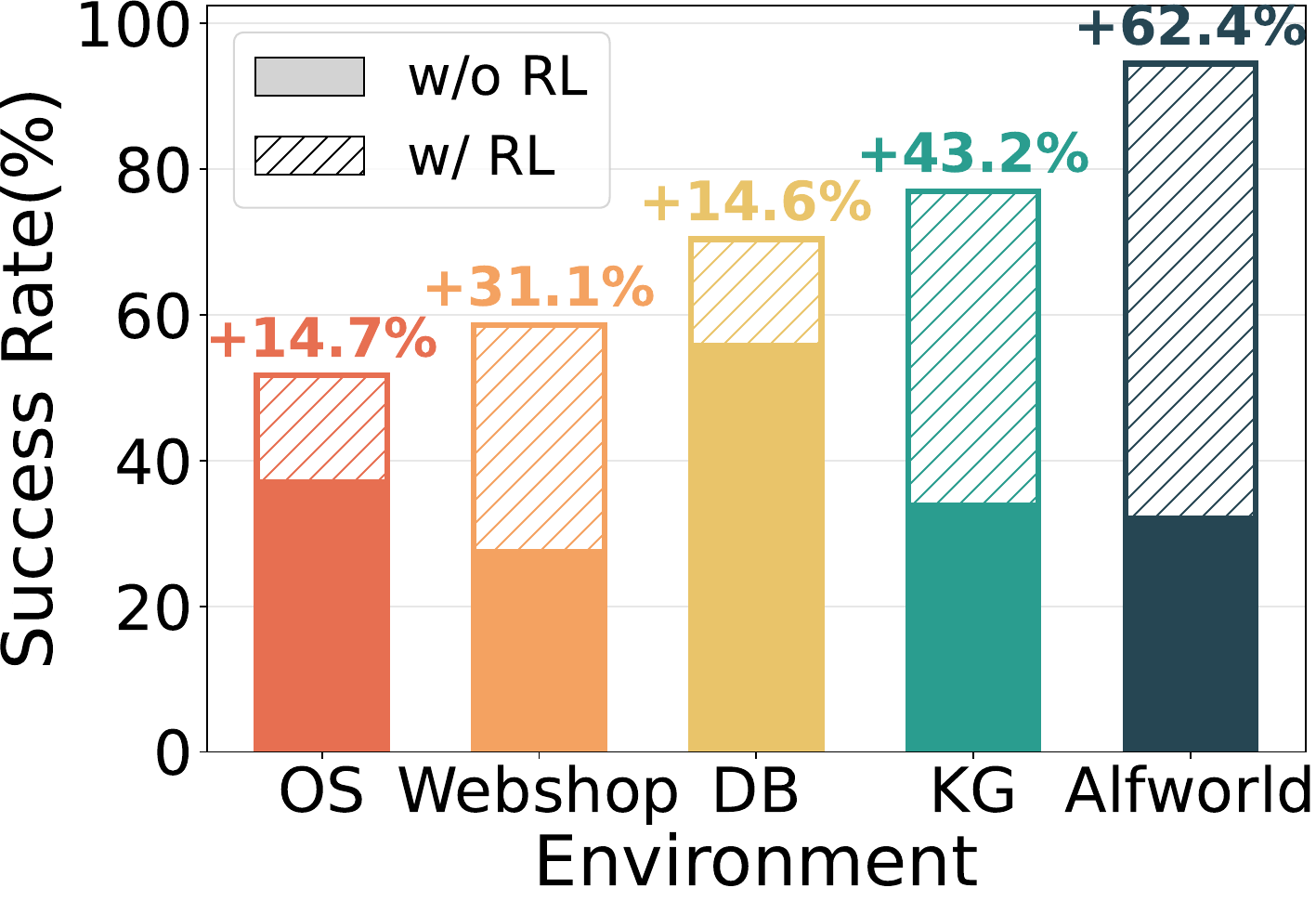}
    \caption{Accuracy gain of \model over the base model in AgentBench, with the hatched area showing the contribution from Reinforcement Learning.}
  \end{subfigure}
  \quad
  \begin{subfigure}[c]{0.45\linewidth}
    \centering
    \includegraphics[width=\linewidth]{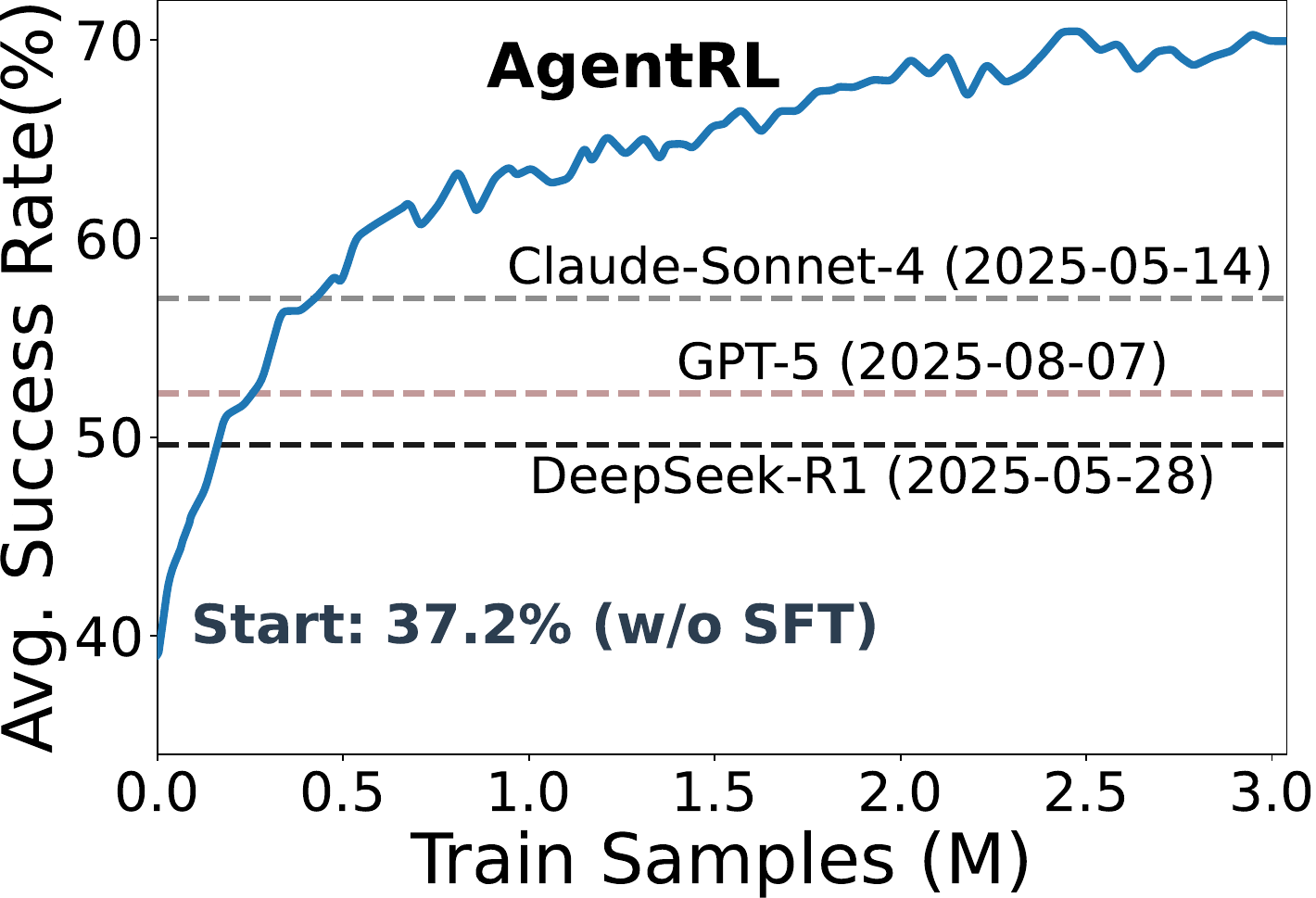}
    \caption{Training progress of \model (32B) on \benchmark without SFT. The solid red line is the average score; dashed lines are baselines.}
    \label{fig:curve}
  \end{subfigure}
  \caption{Overall training result on \benchmark with \model.}
  \label{fig:first}
\end{figure}
\vspace{-5mm}

}

\clearpage
\section{Introduction}
\label{sec:intro}

Reinforcement learning (RL) trains an agent to act by interacting with an environment and optimizing its policy to maximize cumulative rewards.
This principle has been effectively adapted for large language models (LLMs) through reinforcement learning from human feedback (RLHF)~\citep{ouyang2022training,chagpt}, where the LLM itself acts as the agent and its policy is refined based on feedback from a learned reward model. 
This optimization process, typically based on proximal policy optimization (PPO)~\citep{schulman2017proximal}, aligns the model's outputs with desired behaviors.

More recently, reinforcement learning with verifiable rewards (RLVR)~\citep{shao2024deepseekmathpushinglimitsmathematical} has extended RL to reasoning tasks. 
Instead of relying on a learned reward model, RLVR uses automatically verifiable signals, such as correctness checks in math or unit tests in code. 
This shift to objective rewards enables significant simplification of the algorithmic design. 
For example, the group relative policy optimization (GRPO)~\citep{shao2024deepseekmathpushinglimitsmathematical} algorithm further simplifies PPO and improves LLMs' RL training efficiency. 
Recent LLMs leveraging RLVR---e.g., DeepSeek-R1~\citep{deepseekai2025deepseekr1incentivizingreasoningcapability} and T1~\citep{hou2025advancing}---have achieved strong performance in reasoning.

\begin{table*}[t]
\caption{\model vs. other RL frameworks and methods. Interactive Envs: real-time interaction with the environment during training; Heterogeneous Envs: training with diverse environments.}
\label{tab:framework_compare}
\vspace{-2mm}
\centering
\small
\fontsize{8pt}{10pt}\selectfont
\setlength{\tabcolsep}{2pt}
\renewcommand{\arraystretch}{0.9}
\begin{tabular}{@{}p{4.3cm}|cc|ccc@{}} %
\toprule
\multirow{2}{*}{\textbf{Method}} &
\multicolumn{2}{c}{\textbf{Agentic Setting}} &
\multicolumn{3}{c}{\textbf{Infrastructure}} \\
\cmidrule(r){2-3} \cmidrule(r){4-6}
& \textbf{Multi-Turn} & \textbf{Multi-Task}
& \textbf{Full-Async} & 
\textbf{Interative Envs} & \textbf{Heterogeneous Envs}
\\
\midrule
VeRL~\citep{sheng2024hybridflow}        & \xm & \xm & \xm & \xm & \xm \\
OpenRLHF~\citep{hu2024openrlhf}         & \xm & \xm & \xm & \xm & \xm \\
NeMo-Aligner~\citep{shen2024nemo}       & \xm & \xm & \xm & \xm & \xm \\
AReaL~\citep{fu2025areal}               & \cm & \xm & \cm & \xm & \xm \\
\midrule
AgentTuning~\citep{zeng2024agenttuning} & \cm & \cm & \xm & \xm & \xm \\
EasyR1~\citep{zheng2025easyr1}          & \xm & \xm & \xm & \xm & \xm \\
DigiRL~\citep{bai2024digirltraininginthewilddevicecontrol} & \cm & \xm & \xm & \cm & \xm \\
RAGEN~\citep{ragen}                      & \cm & \xm & \xm & \cm & \xm \\
ToolRL~\citep{qian2025toolrl}            & \xm & \xm & \xm & \xm & \xm \\
GiGPO~\citep{feng2025group}              & \cm & \xm & \xm & \cm & \xm \\
ARPO~\citep{lu2025arpo}                  & \cm & \xm & \xm & \cm & \xm \\
\textbf{\model~(ours)}                   & \cm & \cm & \cm & \cm & \cm \\
\bottomrule
\end{tabular}
\vspace{-7mm}
\end{table*}

However, these RL for LLM achievements have been largely limited to \textit{single-turn} settings for a \textit{single task}, where an agent interacts with the given environment only once for feedback~\citep{qi2024webrl,bai2024digirltraininginthewilddevicecontrol,zheng2025deepresearcherscalingdeepresearch,feng2025group,qian2025toolrl,yue2023mammoth}. 
First, to solve agentic tasks with \textit{multi-turn} settings~\citep{openai_deep_research_2025,jin2025searchr1trainingllmsreason,lu2025arpo,feng2025group,lu2025deepdive}, the agent must collect feedback through dynamic interactions with environments~\citep{mind2web,wei2025browsecompsimplechallengingbenchmark}. 
In this case, the LLM is trained as an autonomous agent that performs multi-turn
reasoning, interacts with tools or environments, and adapts its behavior over extended trajectories, that is, the problem of agentic RL. 
Second, building a generalist agent that can handle \textit{diverse tasks} has long been a goal for RL. 
Scaling to heterogeneous multi-task environments in multi-turn settings for agentic RL  requires advances in both LLM training infrastructure and algorithm design. 
Table \ref{tab:framework_compare} lists existing solutions.

In this work, we present a multi-turn, multi-task framework \model to scale agentic RL training. 
\model includes RL infrastructure, environment, and algorithm designs to address the challenges listed in Table \ref{tab:challenges}. 
On the infrastructure side, 
we implement an asynchronous generation-training pipeline that can reduce GPU idle bubbles and improve multi-turn training efficiency.  
On the environment side, we develop a scalable environment deployment infrastructure with a unified function-call based API interface, containerized deployment, and centralized controller to manage the lifecycle of thousands of parallel training episodes. 
To further support heterogeneous environment scaling, we introduce consistent interfaces at the controller level. 
On the algorithm side, we present the cross-policy sampling strategy to encourage model exploration that is negatively impacted by the large state space in the multi-turn setting. 
We also introduce task advantage normalization to mitigate the training instability resulting from the heterogeneity in 
different tasks.

We apply \model on open LLMs---Qwen2.5~\citep{qwen25} and GLM-4-9B~\citep{glm2024chatglm}---across five agentic tasks: ALFWorld, DB, KG, OS, and Webshop \citep{shridhar2020alfworld,webshop,liu2024agentbench}. 
Experiments show that \model
achieves state-of-the-art results, significantly outperforming GPT-5~\citep{gpt5}, Claude-Sonnet-4~\citep{claude4} and DeepSeek-R1~\citep{deepseekai2025deepseekr1incentivizingreasoningcapability} (Figure \ref{fig:first}).  
The single model trained with five tasks together can match the best performance of five models trained separately for individual tasks, while also generalizing into unseen tasks, e.g., BFCL-v3~\citep{patil2025bfcl}. 
Finally, extensive ablations demonstrate that the algorithmic design choices
in \model bring consistent performance benefits. 

The contributions of this work are summarized as follows:

\begin{itemize}[leftmargin=*,itemsep=0pt,parsep=0.2em,topsep=0.2em,partopsep=0.0em]
    
    \item We develop an asynchronous, multi-task framework \model for scalable agentic RL training and robust heterogeneous environment deployment.
    
    \item We design a cross-policy sampling strategy to encourage exploration in multi-turn settings and task advantage normalization to stabilize multi-task RL training.
    
    \item \model achieves state-of-the-art results on various LLM agent tasks, with promising generalization to unseen tasks, demonstrating the potential of building a generalist LLM agent. 
    
\end{itemize}

\hide{

\section{Introduction}
\label{sec:intro}

Recent advances in large language models (LLMs)~\citep{achiam2023gpt, touvron2023llama, touvron2023llama2, glm2024chatglm, zeng2022glm, zhang2022opt, scao2022bloom, team2023gemini} have opened new possibilities for building general-purpose agents capable of reasoning, acting, and interacting across a wide range of tasks. While much of the existing progress has focused on prompt engineering~\citep{yao2023reactsynergizingreasoningacting,wang2023voyageropenendedembodiedagent} or supervised finetuning~\citep{yang2023appagent,zeng2024agenttuning}, there is growing interest in using reinforcement learning (RL) to enable online, multi-turn learning, which allows agents to improve through interaction with dynamic environments. However, despite its promise, online RL for LLM-based agents remains underexplored, and current methods face fundamental challenges in both learning efficiency and efficacy.

While some works have explored a range of RL techniques for training LLM-based agents, such as batch online RL~\citep{qi2024webrl} and offline RL~\citep{bai2024digirltraininginthewilddevicecontrol}, truly online learning, where agents improve through continual interaction with dynamic environments, remains relatively underexplored. This is partly due to the unique challenges posed by online RL in an agentic setting, including: 1) \textit{Limited exploration and continual learning difficulties}, where agents struggle to discover diverse behaviors and retain knowledge over time;
2) \textit{Inefficient and unstable sampling}, stemming from synchronized training pipelines that hinder throughput and amplify optimization variance; 3) \textit{Lack of scalable infrastructure for LLM agent training}. Existing environments are typically designed for static evaluation or scripted control, lacking the throughput, modularity, or fault isolation necessary for efficient RL training. As a result, it remains difficult to train and debug agent policies in a realistic, scalable, and reproducible fashion.

In this work, we present a new training methodology and infrastructure for scalable, multi-task RL with LLM agents, as depicted in Figure~\ref{fig:agentrl-fig}. Our approach departs from traditional synchronized training pipelines by introducing a fully asynchronous architecture, in which agent rollouts and training are decoupled for maximal hardware efficiency and training stability. We redesign the agent-environment interface, incorporating a unified function-call API, a high-performance controller, and support for containerized execution. This infrastructure enables fast, consistent, and scalable rollout generation, significantly lowering the engineering burden of deploying RL at scale.

To overcome challenges in balancing exploration and exploitation, we propose a cross-policy sampling strategy that leverages a pool of partially desynchronized models, enabling more diverse action trajectories and better sample efficiency over time. Additionally, we mitigate inter-task interference in multi-task training via a task advantage normalization mechanism.

We validate our framework on \benchmark, demonstrating substantial improvements in overall task performance. Our findings suggest that both algorithmic diversity and infrastructure design are critical to unlocking the full potential of RL for general-purpose LLM agents. Our main results are depicted in Figure \ref{fig:first}.
Our key contribution is summarized as follows:

\begin{itemize}[leftmargin=*,itemsep=0pt,parsep=0.2em,topsep=0.2em,partopsep=0.0em]
    
    \item \textbf{An asynchronous, scalable framework for training and environment deployment.} We design a fully asynchronous RL system that alleviates major efficiency bottlenecks in existing pipelines. By reconstructing \agentbench's environment backend with a unified API and containerized execution, our framework achieves higher throughput, better GPU utilization, and supports efficient scaling and integration for LLM agent training.
    
    \item \textbf{Cross sampling strategy and task advantage normalization method to boost and stablize RL training.} We propose a cross sampling method that performs steps from a pool of partially desynchronized models to leverage model exploration diversity. And to stablize multi-task RL training, we also introduce an advantage normalization in task level.
    
    \item \textbf{State-of-the-art results on multi-task LLM agent benchmarks.} Leveraging our redesigned deployment infrastructure based on \agentbench, we demonstrate that our method achieves new SOTA performance across \benchmark environments. Our architecture improves sample efficiency, generalization, and robustness in complex real-world interaction settings, benefiting from consistent environment handling and scalable training.
\end{itemize}

}

\section{The Agentic RL Problem and its Challenges}

\begin{table}[t]
\centering
\small
\caption{Challenges in agentic RL compared to single-turn RL}
\renewcommand{\arraystretch}{1.1}
\setlength{\tabcolsep}{5pt}
\begin{tabular}{@{}m{2cm} | m{6cm} | m{5cm}@{}}
\toprule
& \textbf{Infrastructure} & \textbf{Algorithm} \\
\midrule
\textbf{Single-Turn} 
& synchronous rollouts 
& stable and scalable training \\
\midrule
\textbf{Multi-Turn} 
& compute inefficiency in synchronous rollouts, requiring asynchronous training; 
difficulty in scaling interactive homogeneous environments 
& multi-turn tasks demand greater exploration due to larger state spaces, but exploration declines during training \\
\midrule
\textbf{Multi-Task} 
& difficulty in unifying heterogeneous environments 
& performance drop from task interference and lack of generalization \\
\bottomrule
\end{tabular}
\vspace{-0.5em}
\label{tab:challenges}
\end{table}

The shift from single-turn to multi-turn defines the problem of agentic RL, where the LLM acts as an autonomous agent that performs multi-turn reasoning, interacts with tools or environments, and adapts its behavior over extended trajectories. 
Formally, this can be formulated as a Markov Decision Process(MDP)~\citep{puterman2014markov}, a tuple $(\mathcal{S},\mathcal{A},P,r,\rho)$, 
where $\mathcal{S}$ is the state set, 
$\mathcal{A}$  the action set, 
$P$   the state-transition probability, 
$r$   the reward function, 
and 
$\rho$   the initial state distribution. 
In a single-step case, $P$ is trivial and the problem reduces to a multi-armed bandit. 
In contrast, multi-step MDPs involve non-trivial state evolution over multiple transitions.
The definition is listed in Appendix~\ref{sec:pre}.

Moreover, most LLM agents have focused on training a separate policy  for each individual task~\citep{zheng2025deepresearcherscalingdeepresearch,feng2025group,qian2025toolrl}. 
That means multiple LLMs have to be trained, one for each environment or task, respectively. 
How to build a generalist agent that can handle diverse tasks remains largely unexplored. 
Table~\ref{tab:challenges} summarizes the challenges that go beyond single-turn RL.

\textbf{Infrastructure Challenges in Multi-Turn RL.}  
In the single-turn setting,  RL is often run in a synchronous way with an interleaved generation-training pipeline~\citep{hu2024openrlhf,sheng2024hybridflow}. 
For agentic tasks, generating long trajectories and frequent interactions with the environment is slow, time-consuming, and highly variable compared to single-turn scenarios. 
As a result, GPUs that handle short trajectories have to stay idle to wait for the generation completion of long trajectories. 
The imbalance significantly reduces training efficiency and prevents RL scaling, thus requiring an asynchronous RL training framework. 

On the environment side, multi-turn training requires rollouts to run in an interactive environment, which places high demands on the concurrent deployment and management of a large number of homogeneous environments. 

\textbf{Algorithm Challenges in Multi-Turn RL.}  
On the algorithm side,  most existing sampling strategies are designed for single-turn settings. 
Improving exploration and sampling efficiency in multi-turn scenarios is therefore critical for agentic RL training.

\textbf{Infrastructure Challenges in Multi-Task  RL.}  
By definition, multi-task RL requires an architecture that can manage diverse environments. 
One major challenge lies in the differences in environment interfaces, state-action representations, and computational demands. 
Effective and scalable integration of these environments is essential for scaling agentic training efficiently across diverse tasks.

\textbf{Algorithm Challenges in Multi-Task  RL.}  
Most existing RL approaches focus on training a single agent task~\citep{jin2025searchr1trainingllmsreason,qian2025toolrl,feng2025group}. 
Thus, developing effective methods for jointly optimizing multiple agent tasks while ensuring training stability remains an open challenge.

\hide{

\subsection{Problem Formulation}

Reinforcement Learning (RL) is a paradigm where an agent learns to act by interacting with an environment and optimizing its policy to maximize cumulative rewards.
This principle was powerfully adapted for LLMs in the RLHF~\citep{ouyang2022training} framework. In this approach, the LLM itself acts as the agent, and its policy is refined based on feedback from a reward model.
This optimization process, typically using Proximal Policy Optimization (PPO)~\citep{schulman2017proximal}, aligns the model's outputs with desired behaviors.

While the RLHF paradigm is mainly used to solve alignment tasks,
a recent paradigm, Reinforcement Learning with Verifiable Rewards(RLVR)~\citep{deepseekai2025deepseekr1incentivizingreasoningcapability}, further extends RL methods to reasoning tasks.
RLVR replaces the learned reward model with automatically verifiable signals, such as correctness checks in mathematics or unit tests in code generation. 
This shift to objective rewards enables significant algorithmic simplification, exemplified by 
Group Relative Policy Optimization (GRPO)~\citep{shao2024deepseekmathpushinglimitsmathematical} algorithm further simplifies PPO, boosting RL training efficiency. Recent works leveraging the GRPO algorithm and RLVR paradigm have demonstrated significant success in reasoning domains such as mathematics and code generation.~\citep{jaech2024openai,shao2024deepseekmathpushinglimitsmathematical,hou2025advancing,deepseekai2025deepseekr1incentivizingreasoningcapability}.

However, these achievements have been largely confined to single-turn settings, where an agent interacts with the environment only once for feedback. To solve real-world tasks, recent works apply the RLVR paradigm to more complex interactive settings~\citep{openai_deep_research_2025,jin2025searchr1trainingllmsreason,lu2025arpo,feng2025group}, which formulate multi-turn settings. In these formulations, LLMs are treated as autonomous agents that perform multi-turn reasoning, interact with tools or environments, and adapt their behavior across extended trajectories.

To date, most LLM-based agent efforts have focused on training a single policy model for a single task~\citep{zheng2025deepresearcherscalingdeepresearch,feng2025group,qian2025toolrl}, which has limited ability in different scenarios. Therefore, to build more generalized agents, we further extend to a multi-task framework designed to train a single, versatile policy capable of handling diverse tasks.

These extensions from single-turn to multi-turn and multi-task settings give rise to what we term as the agentic RL paradigm, where the LLM functions as a goal-directed, interactive agent. The problem can be formulated into a Markov Decision Process(MDP), a tuple $(\mathcal{S},\mathcal{A},P,r,\rho)$, where $\mathcal{S}$ is the state set, $\mathcal{A}$ the action set, $P$ the state-transition probability, $r$ the reward function, and $\rho$ the initial state distribution. In the single-step case $P$ is trivial and the problem reduces to a multi-armed bandit, whereas multi-step MDPs involve nontrivial state evolution over multiple transitions. The full definition can be found in Appendix~\ref{sec:pre}.

Table~\ref{tab:challenges} summarizes the differences and challenges between single-turn, multi-turn, and multi-task RL. 
The comparison covers both infrastructure and algorithmic perspectives. 

========

\section{Main Challenges}

\begin{table}[t]
\centering
\small
\caption{Challenges in agentic RL compared to single-turn RL}
\renewcommand{\arraystretch}{1.1}
\setlength{\tabcolsep}{5pt}
\begin{tabular}{@{}m{2cm} | m{6cm} | m{5cm}@{}}
\toprule
& \textbf{Infrastructure} & \textbf{Algorithm} \\
\midrule
\textbf{Single-Turn} 
& synchronous rollouts 
& stable and scalable training \\
\midrule
\textbf{Multi-Turn} 
& compute inefficiency in synchronous rollouts, requiring asynchronous training; 
difficulty in scaling interactive homogeneous environments 
& multi-turn tasks demand greater exploration due to larger state spaces, but exploration declines during training \\
\midrule
\textbf{Multi-Task} 
& difficulty in unifying heterogeneous environments 
& performance drop from task interference and lack of generalization \\
\bottomrule
\end{tabular}
\vspace{-0.5em}
\label{tab:challenges}
\end{table}

Table~\ref{tab:challenges} summarizes the primary challenges faced by agentic RL, 
categorized along two dimensions: \emph{multi-turn} vs. \emph{multi-task} settings, and 
\emph{infrastructure} vs. \emph{algorithmic} perspectives. We elaborate on each category below.

\subsection{Scaling End-to-End Agentic Multi-Turn Reinforcement Learning}

\paragraph{Infrastructure.}  
Training end-to-end agentic multi-turn RL imposes substantial infrastructure demands. In typical synchronous RL pipelines, episodes of shorter length must wait for longer ones to finish, leading to prolonged idle time and significant resource waste. In addition, multi-turn training requires rollouts to be executed in real interactive environments, which places a high demand on concurrently deploying and managing large numbers of environments.

\textit{Typical synchronous reinforcement learning (RL) runs in an interleaved generation-training pipeline. Agent generation is more time-consuming and exhibit greater variance compared to chatting, caused by the longer trajectories and continuous interaction with the environment. As a result, the GPUs for most short trajectory generation are idle in most time, waiting for few super long trajectory generation to finish. The imbalance generation significantly slows down training efficiency and prevent RL scaling. In addition, multi-turn training requires rollouts to be executed in real interactive environments, which places a high demand on concurrently deploying and managing large numbers of environments.}

\paragraph{Algorithm.}  
From the algorithmic side, the efficiency and effectiveness of rollouts are critical for RL training. However, LLMs tend to become less exploratory as training progresses, which can limit the ultimate performance of agentic RL models.

\textit{From the algorithmic side, exploration and sampling efficiency matter for RL training. Previous sampling strategies focus on single-turn sampling while few works explores how to conduct more effective sampling strategy for better search efficiency.}

\subsection{Unifying Multi-Task Reinforcement Learning Training}

\paragraph{Infrastructure.}  
Multi-task RL necessitates an architecture capable of orchestrating heterogeneous environments in a unified manner. Differences in environment interfaces, state-action representations, and computational requirements make integration nontrivial, and the lack of such a framework hinders the ability to scale training seamlessly across diverse tasks.

\paragraph{Algorithm.}  
Compared with single-task settings, jointly optimizing across multiple tasks frequently leads to interference between task objectives, usually causing a degradation in single-task performance. Preserving single-task performance while simultaneously training across multiple tasks remains a non-trivial challenge.

\textit{Previous works mostly conduct RL training on a single agent task. It is more practical to jointly optimize multiple agent tasks to unify different capabilities into one model. It remains under-explored how to conduct multi-task agent RL training and further activate the generalization to unseen tasks.}

}

\section{The \model Framework}

\begin{figure}[t]
    \centering
    \includegraphics[width=.9\linewidth]{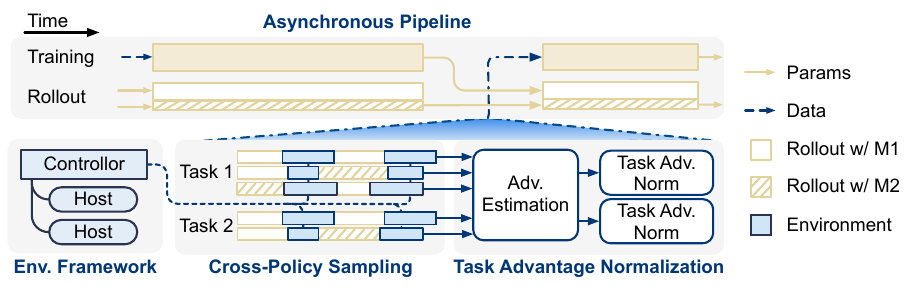}
    \caption{An overview of \model. 
    Top: asynchronous training and rollout flows. 
    Bottom: the environment framework where a controller manages multiple workers to provide environments, and the rollout details, including cross-policy sampling and task advantage normalization. }
    \label{fig:agentrl-fig}
\end{figure}

In this work, we develop an agentic RL framework---\model---to support multi-turn and multi-task RL training, as shown in Figure \ref{fig:agentrl-fig}. 
\model implements asynchronous training and environment deployment to improve efficiency in multi-turn and multi-task settings.  
It also introduces cross-policy sampling and task advantage normalization to stabilize the RL training. 
Together, these technical designs and implementations address the challenges outlined in Table \ref{tab:challenges}, and thus enable the generalist agent training by scaling multiple environments. 

\subsection{Multi-Turn Agentic RL}
\label{sec:multi-turn}

\textbf{Asynchronous Training Framework.}
To overcome the efficiency bottlenecks of synchronous batching, we introduce an asynchronous rollout-training strategy based on coroutine scheduling.
The rollout engine runs in a dedicated resource group and executes asynchronously with training. 
The training module continuously pulls available data from the rollout engine after each update, without waiting for an entire batch of rollouts to finish. 
In addition, it accepts a dynamic batch size that fluctuates within a certain range.
This design enables the scheduler to fill idle GPU slots with available coroutines, reducing pipeline bubbles and improving overall throughput.

\begin{figure}[htbp]
    \centering
    \includegraphics[width=0.9\linewidth]{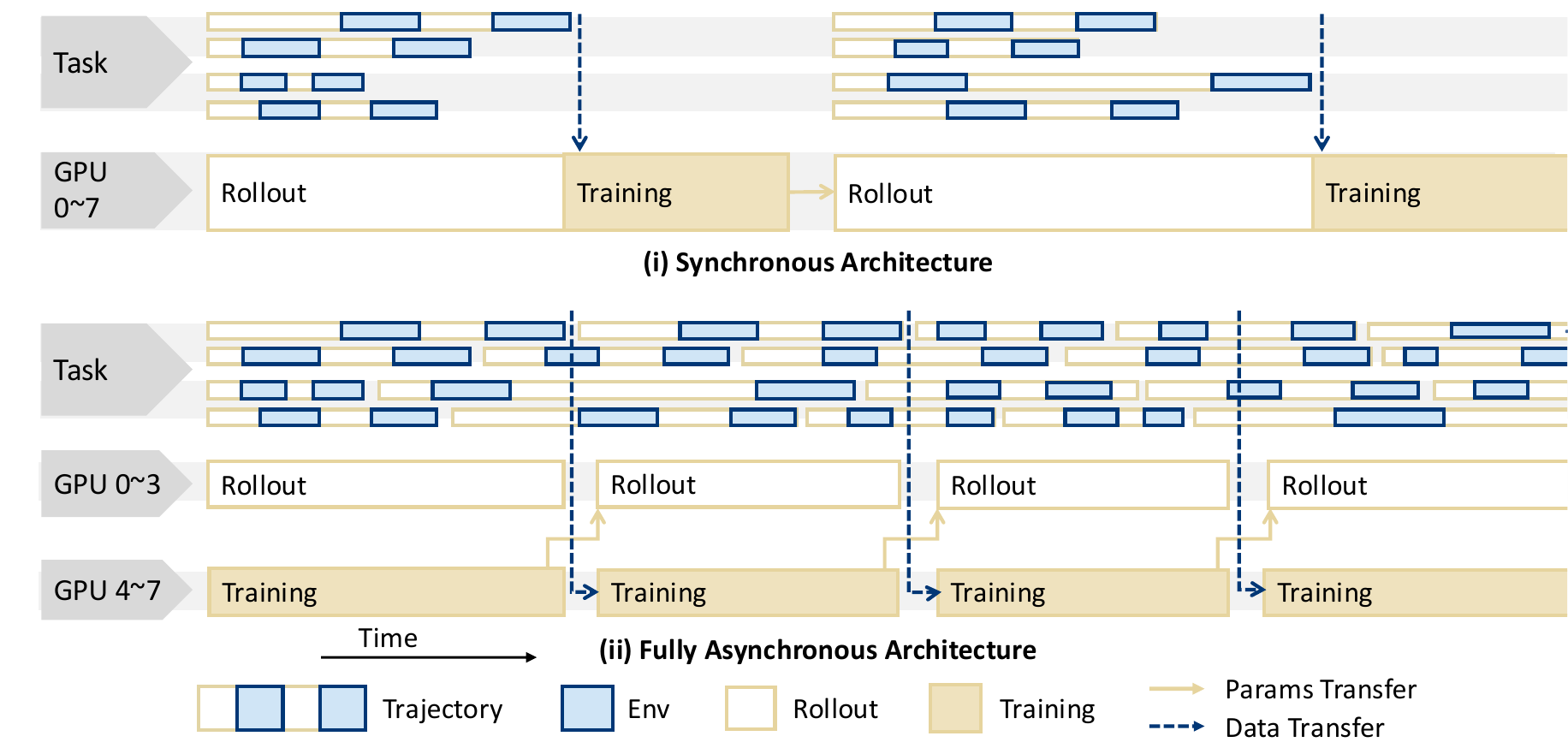}
    \vspace{-2mm}
    \caption{Synchronous vs. Asynchronous Training. 
    The asynchronous design improves efficiency by separating data rollout and model training on different resource groups.}
    \label{fig:fully-async-arch}
\end{figure}

\begin{wrapfigure}{r}{0.5\textwidth}
    \centering
    \includegraphics[width=0.9\linewidth]{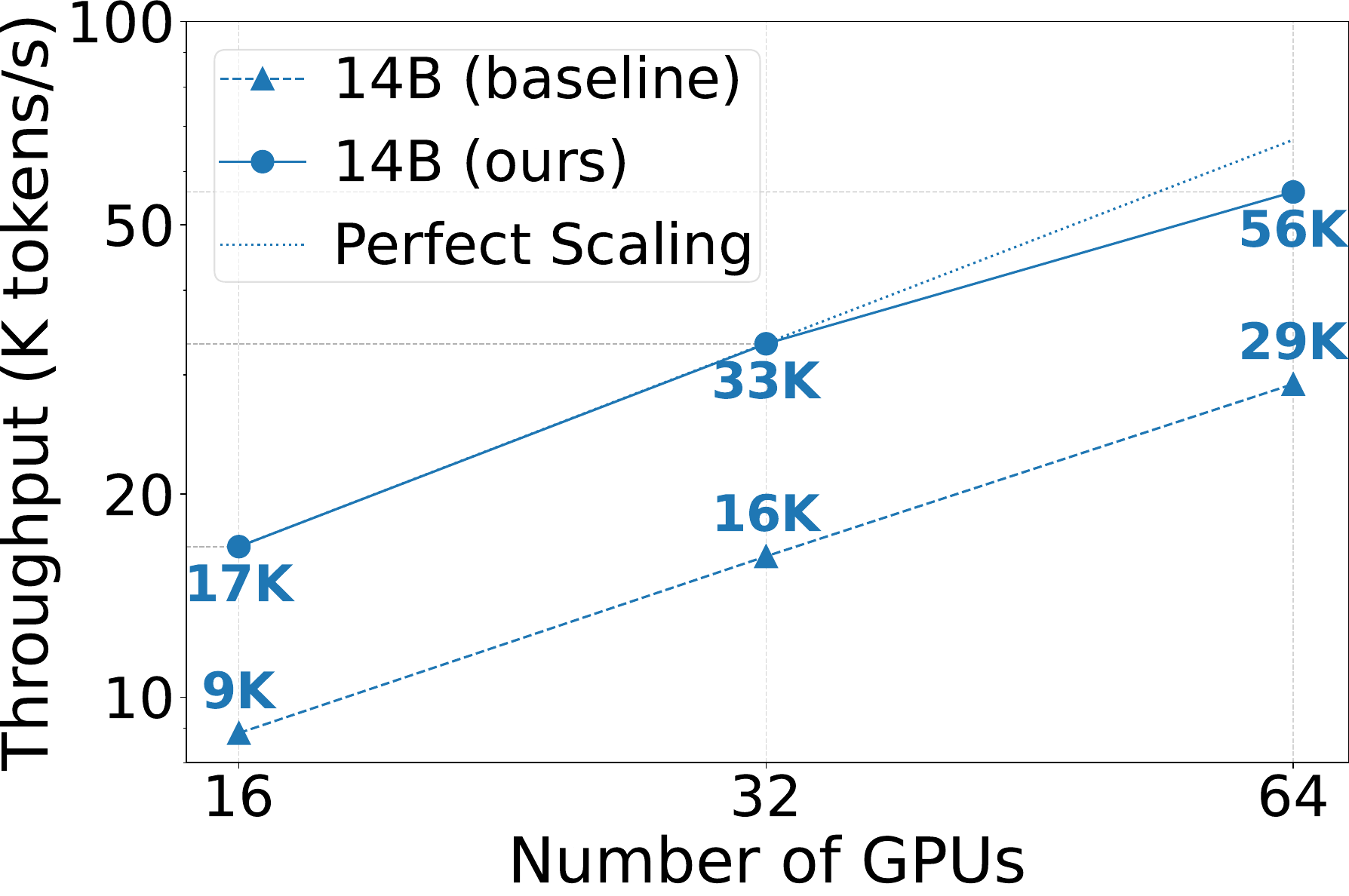}
    \caption{Throughput of \model vs. the synchronous baseline for 14B parameter (Qwen2.5) models on Webshop (log-scale for both axes).}
    \label{fig:scaling}
    \vspace{-5mm}
\end{wrapfigure}

As illustrated in Figure~\ref{fig:fully-async-arch},  rollout and training are decoupled---they run concurrently and communicate asynchronously. 
This enables efficient hardware scheduling, as shown in Figure~\ref{fig:scaling}, where the asynchronous pipeline in \model brings significant throughput gains over the synchronous one.

To avoid off-policy bias of the rollout engine, we set a maximum size of the data queue and  enforce all trajectories to be moved to the training  engine at each step. 
In doing so, all trajectories are kept as up-to-date as possible with the latest policy, which later experiments suggest to be acceptable.

\begin{figure}[t]
  \centering
  \includegraphics[width=0.99\linewidth]{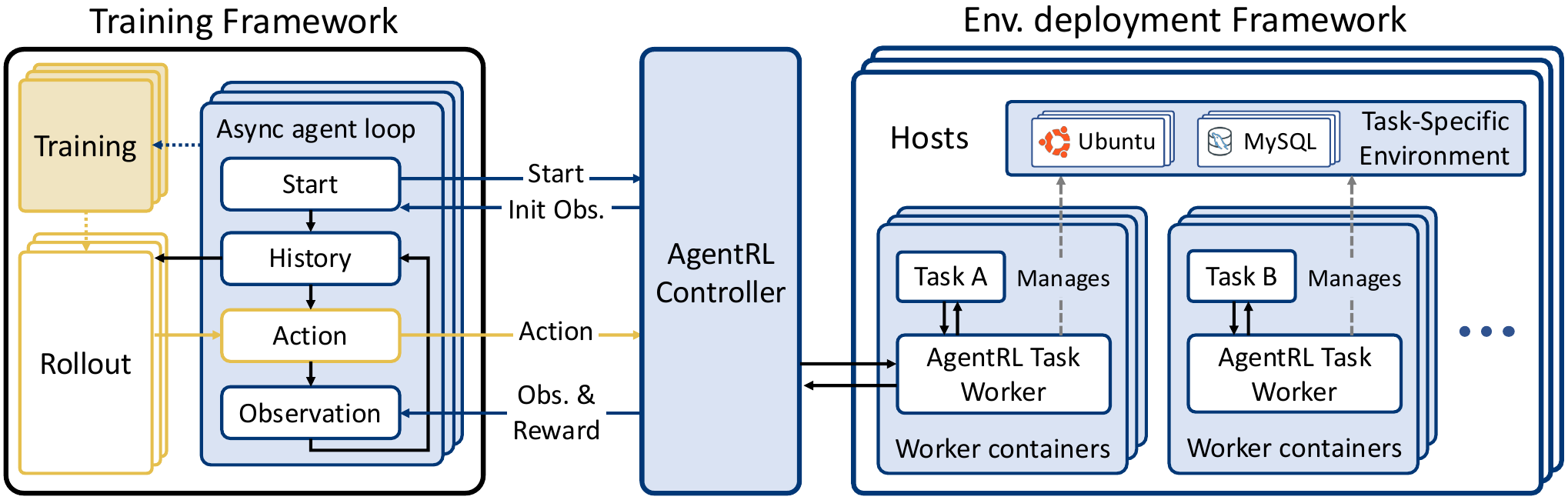}
  \hfill
  \caption{The \model training pipeline, decoupled into a Training Framework and an Environment Deployment Framework, organized by a central \model Controller. 
  The Training Framework is responsible for policy rollouts and updates, while the Environment Deployment Framework manages scalable, containerized task environments that provide feedback.}
  \label{fig:overall-arch}
  \vspace{-6mm}
\end{figure}

\textbf{Scalable Agentic Environment Infrastructure.}
To enable large-scale agentic RL, we develop a scalable environment deployment infrastructure, shown in Figure \ref{fig:overall-arch}.
It includes the following components: 
\textit{1. Function-call based environment interface.} 
To simplify environment interactions, we introduce a unified, function-call based API. 
This replaces complex custom action formats and thus enables centralized management and monitoring.
\textit{2. Containerized deployment.} 
Each task environment is containerized as an isolated execution unit. 
This design improves resource allocation, isolates faults between concurrent sessions, and supports seamless deployment on diverse hardware. 
\textit{3. Centralized high-performance controller.} 
A central controller,
acts as the global orchestrator for the training engine. 
It is optimized for high-concurrency workloads and manages the lifecycle of thousands of parallel training episodes.

\begin{wrapfigure}{r}{0.5\textwidth}
    \centering
  \includegraphics[width=1.0\linewidth]{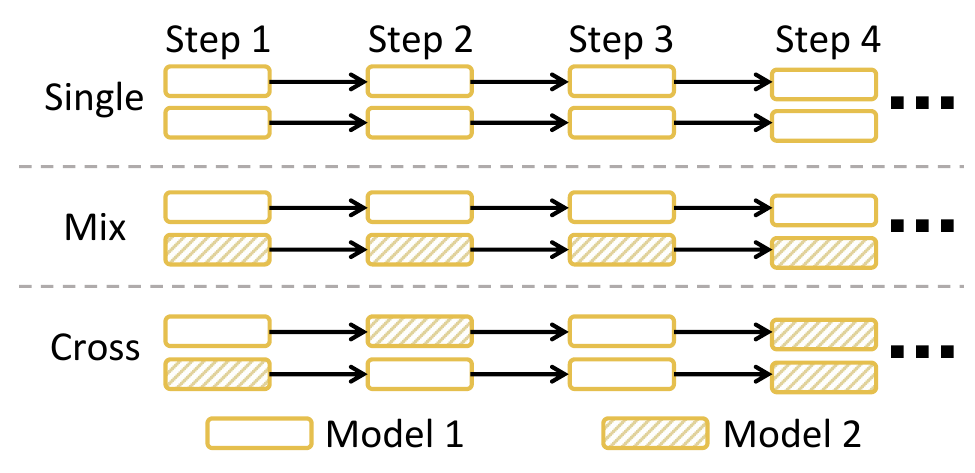}
  \caption{Different rollout strategies. 
  In \textit{single} model generation, all steps of all traces are generated by the same model. 
  In \textit{mix} mode, half of the samples are generated by each model. 
  In \textit{cross-policy} mode, all samples are generated with cross-policy sampling strategy.}
  \label{fig:rollouts}
  \vspace{-5mm}
\end{wrapfigure}
\textbf{Cross-Policy Sampling Strategy.}

During RL training, model exploration typically declines over time. 
This problem becomes more severe in the multi-turn setting with large state spaces. 
In addition, \emph{model collapse} ~\citep{shumailov2024ai} has been reported, where repeated training on self-generated data leads to degraded capability and reduced variance.

To overcome this issue, we propose a cross-policy sampling strategy (see Figure \ref{fig:rollouts}), where multiple LLMs are used to generate actions with a single trajectory. 
The goal of aggregating  data from different models is to increase the diversity of the candidate pool while preserving overall quality. 
Specifically, cross-policy sampling constructs trajectories by allowing actions at each step to be randomly drawn from the pool of available models, rather than committing to a single model. 

Its advantage lies in that the language component of each state is still constrained to remain valid, while the expanded sampling enlarges the coverage of language states that can reach successful outcomes in the environment. 
By exploring paths that would not appear under any single model, cross-policy sampling increases the likelihood of visiting goal-relevant states without drifting into incoherent or invalid linguistic regions. 
Details can be found in Appendix~\ref{sec:formal-cs}.

During RL training, it is hard to incorporate models with different architectures in the pipeline. 
Instead, we let the model do cross-policy sampling with its early version. Specifically, we mark a set of rollout engines as stale engines; these engines update parameters every multiple steps instead of one step. 
Early experiments verified the effect of the cross-policy sampling strategy (see Section \ref{sec:cross-exp}).

\subsection{Multi-Task Agentic RL}

\textbf{Heterogeneous Environment Deployment.}
Multi-task RL requires the environment deployment framework to generalize beyond a single task or environment. 
To host, schedule, and monitor heterogeneous environments under the same infrastructure without incurring additional integration cost, we propose to  expose consistent interfaces at both the worker and controller levels.  
This supports \model to scale the task (environment) set in size and diversity gracefully. 

We have two complementary designs: 
On the \textit{environment} side, we unify the worker API across all tasks, such that each task can be instantiated and managed using an identical set of lifecycle operations. 
On the \textit{training} side (Figure \ref{fig:overall-arch}), the controller provides a single gateway API to the RL engine, abstracting away task heterogeneity and exposing multi-task execution as a transparent extension of the single-task case.

\textbf{Task Advantage Normalization.}
In multi-task RL, agentic tasks  often differ substantially in difficulty, sequence length, and sampling efficiency. 
Such heterogeneity can cause standard RL algorithms to learn at very different rates across tasks. 
Consequently, one task may exhibit clear reward improvements where another shows negligible progress, leading to training instability and performance imbalance.

For an LLM-based policy, each high-level action $a_t$ consists of multiple tokens
$\{y_{t,k}\}_{k=1}^{L_t}$. 
We compute token-level advantage estimates $\hat{A}_{i,s,g,t,k}$ for each token occurrence, 
where $i$ denotes the task index, 
$s$ the sample index within the task, 
$g$ the trajectory index within the group, 
$t$ the environment step, 
and $k$ the token position within $a_t$.

Let
$
    \mathcal{A}_i^{\mathrm{tok}} = \left\{ \hat{A}_{i,s,g,t,k} \ \middle|\ 
    1 \le s \le S_i,\ 
    1 \le g \le K_{i,s},\ 
    1 \le t \le T_{i,s,g},\
    1 \le k \le L_{i,s,g,t} 
    \right\}
$ denote the set of token-level advantages for all tokens in 
the current batch of task $i$, 
where $S_i$ is the number of samples, $K_{i,s}$ the number of trajectories per sample, 
$T_{i,s,g}$ the number of env steps in trajectory $\tau_{i,s,g}$, 
and $L_{i,s,g,t}$ the number of tokens in action $a_t$.

We normalize each token's advantage within its task batch as:
\begin{equation}
    \tilde{A}_{i,s,g,t,k} = 
    \frac{\hat{A}_{i,s,g,t,k} - \mu_i}
         {\sigma_i},
\end{equation}
where $\mu_i = \mathrm{mean}(\mathcal{A}_i^{\mathrm{tok}})$ and 
$\sigma_i = \mathrm{std}(\mathcal{A}_i^{\mathrm{tok}})$.
This ensures that, for each task $i$, the distribution of token-level advantages in a batch has zero mean and unit variance, helping to reduce inter-task variance and stabilize multi-task optimization.

\section{Experiments}
\label{sec:exp}

\begin{table*}[t]
\caption{Main results (task success rate). 
Average and standard deviation of four repeats on each task are reported. 
The `*' indicates reward results directly extracted from the original papers. 
}
\vspace{-1mm}
\label{tab:main}
\centering
\small
\footnotesize
\begin{threeparttable}
\renewcommand{\arraystretch}{1} %
\begin{tabular*}{\columnwidth}{@{}l @{\extracolsep{\fill}} S[table-format=2.1,detect-weight] S[table-format=2.1,detect-weight] S[table-format=2.1,detect-weight] S[table-format=2.1,detect-weight] S[table-format=2.1,detect-weight] S[table-format=2.1,detect-weight] @{}}
\toprule
\textbf{Model} & {\textbf{ALFWorld}} & {\textbf{DB}} & {\textbf{KG}} & {\textbf{OS}} & {\textbf{Webshop}} & {\textbf{AVG}} \\
\midrule
\multicolumn{7}{c}{\textit{API LLMs (Prompting)}} \\
\midrule
Claude-Sonnet-3.7 (2025-02-19) & 61.1\err{3.0} & 68.5\err{0.8} & 59.8\err{1.0} & 36.5\err{4.1} & 40.1\err{1.5} & 53.2 \\
Claude-Sonnet-3.7 Thinking (2025-02-19) & 54.1\err{3.0} & 68.4\err{0.3} & 38.2\err{2.2} & 53.1\err{1.8} & 36.0\err{1.7} & 50.0 \\
Claude-Sonnet-4 (2025-05-14) & 73.6\err{2.6} & 70.1\err{0.7} & 63.4\err{1.7} & 45.3\err{2.8} & 34.6\err{1.6} & 57.4 \\
Claude-Sonnet-4 Thinking (2025-05-14) & 69.0\err{3.2} & 68.4\err{1.0} & 64.4\err{1.9} & 51.0\err{2.3} & 38.3\err{2.8} & 58.2 \\
GPT-4o (2024-11-20) & 28.3\err{2.8} & 54.3\err{2.2} & 49.3\err{2.7} & 38.5\err{3.2} & 27.8\err{2.2} & 39.6 \\
o3-mini (2025-01-31) & 28.4\err{1.3} & 56.5\err{0.5} & 51.8\err{0.9} & 35.1\err{1.7} & 32.7\err{1.5} & 40.9 \\
o4-mini (2025-04-16) & 32.6\err{1.8} & 63.4\err{0.3} & 32.4\err{3.0} & 41.8\err{1.0} & 28.5\err{1.8} & 39.7 \\
GPT-5 (2025-08-07) & 65.4\err{2.0} & 63.2\err{0.7} & 64.1\err{1.8} & 34.5 \err{1.0} & 33.7 \err{2.6} & 52.2 \\

\midrule
\multicolumn{7}{c}{\textit{Open LLMs (Prompting)}} \\
\midrule
DeepSeek-V3 (2025-03-24) & 31.9\err{2.0} & 58.4\err{1.2} & 14.0\err{2.0} & 53.0\err{1.0} & 23.4\err{2.5} & 36.1 \\
DeepSeek-R1 (2025-05-28) & 51.4\err{4.1} & 60.4\err{0.5} & 50.2\err{2.7} &  53.6\err{1.0} & 31.0\err{1.6} & 49.3 \\
Qwen2.5-14B-Instruct & 8.7\err{3.1} & 48.4\err{2.2} & 35.3\err{3.0} & 26.0\err{3.1} & 17.6\err{1.0} & 27.2 \\
Qwen2.5-32B-Instruct & 32.1\err{3.9} & 55.8\err{0.6} & 33.8\err{1.5} & 37.0\err{1.5} & 27.5\err{2.3} & 37.2 \\
Qwen2.5-72B-Instruct & 47.5\err{3.3} & 45.3\err{0.9} & 26.5\err{3.1} & 49.5\err{3.5} & 35.4\err{2.7} & 40.8 \\
\midrule
\multicolumn{7}{c}{\textit{Open LLMs (Agent Training)}} \\
\midrule
Hephaestus-8B-Base & 30.0 & 32.3 & 16.0 & 20.8 & 60.5$^{*}$ & 31.9 \\
Hephaestus-8B-IFT & 46.0 & 29.7 & 21.2 & 20.8 & 63.9$^{*}$ & 36.3 \\
AgentLM-7B & 84.0 & 30.6 & 18.1 & 17.4 & 63.6$^{*}$ & 42.7 \\
AgentLM-13B & 76.0 & 33.7 & 26.8 & 18.1 & 70.8$^{*}$ & 45.1 \\
AgentLM-70B & 86.0 & 37.7 & 47.0 & 21.5 & 64.9$^{*}$ & 51.4 \\
\midrule
\multicolumn{7}{@{}l}{\model} \\
\quad w/ Qwen2.5-3B-Instruct   & 92.4\err{0.5} & 60.0\err{1.1} & 55.0\err{2.0} & 40.5\err{0.9} & 52.1\err{0.9} & \bfseries 60.0 \\
\quad w/ Qwen2.5-7B-Instruct   & 91.5$^{\dagger}$\err{0.9} & 63.7\err{0.5} & 57.8\err{2.3} & 40.8\err{1.2} & 56.1\err{0.6} & \bfseries 62.0 \\
\quad w/ Qwen2.5-14B-Instruct   & 91.5\err{0.9} & 72.2\err{0.9} & 72.8\err{1.8} & 43.6\err{1.9} & 58.5\err{1.2} & \bfseries 67.7 \\
\quad w/ Qwen2.5-32B-Instruct  &  94.5 \err{0.5}&  70.4 \err{0.5} & 77.0 \err{1.2} & 51.7\err{1.8} & 58.6 \err{0.9} & \bfseries 70.4 \\
\quad w/ GLM-4-9B-0414  &  93.3 \err{0.5} & 66.9 \err{0.4} &  75.7 \err{1.8} & 33.2 \err{1.7} & 55.9 \err{1.9}& \bfseries 65.0 \\
\bottomrule
\end{tabular*}
\begin{tablenotes}
\footnotesize
\item[$^\dagger$] We provide a one-shot demonstration for Qwen2.5-7B-Instruct in ALFWorld evaluation, as it fails to generate valid tool call format in the environment.
\end{tablenotes}
\end{threeparttable}
\end{table*}

\hide{

\begin{table*}[t]
\caption{Main results on \benchmark. Models are sorted by their average score. We use success rate as the metric. We report average and standard deviation of 4 repeat on each task. The `*' indicates reward scores for Webshop directly reported from the original papers. The best and second-best results in each column are shown in bold and \underline{underlined}, respectively.}
\vspace{-1mm}
\label{tab:main}
\centering
\small
\footnotesize
\begin{threeparttable}
\renewcommand{\arraystretch}{1} %
\begin{tabular*}{\columnwidth}{@{}l @{\extracolsep{\fill}} S[table-format=2.1,detect-weight] S[table-format=2.1,detect-weight] S[table-format=2.1,detect-weight] S[table-format=2.1,detect-weight] S[table-format=2.1,detect-weight] S[table-format=2.1,detect-weight] @{}}
\toprule
\textbf{Model} & {\textbf{ALFWorld}} & {\textbf{DB}} & {\textbf{KG}} & {\textbf{OS}} & {\textbf{Webshop}} & {\textbf{AVG}} \\
\midrule
\multicolumn{7}{c}{\textit{API LLMs (Prompting)}} \\
\midrule
DeepSeek-V3 & 31.9\err{2.0} & 58.4\err{1.2} & 14.0\err{2.0} & 53.0\err{1.0} & 23.4\err{2.5} & 36.1 \\
GPT-4o & 28.3\err{2.8} & 54.3\err{2.2} & 49.3\err{2.7} & 38.5\err{3.2} & 27.8\err{2.2} & 39.6 \\
o4-mini & 32.6\err{1.8} & 63.4\err{0.3} & 32.4\err{3.0} & 41.8\err{1.0} & 28.5\err{1.8} & 39.7 \\
o3-mini & 28.4\err{1.3} & 56.5\err{0.5} & 51.8\err{0.9} & 35.1\err{1.7} & 32.7\err{1.5} & 40.9 \\
DeepSeek-R1 & 51.4\err{4.1} & 60.4\err{0.5} & 50.2\err{2.7} & \bfseries 53.6\err{1.0} & 31.0\err{1.6} & 49.3 \\
Claude-3.7-sonnet-thinking & 54.1\err{3.0} & 68.4\err{0.3} & 38.2\err{2.2} & {53.1\err{1.8}} & 36.0\err{1.7} & 50.0 \\
GPT-5 & 65.4\err{2.0} & 63.2\err{0.7} & 64.1\err{1.8} & 34.5 \err{1.0} & 33.7 \err{2.6} & 52.2 \\
Claude-3.7-sonnet & 61.1\err{3.0} & 68.5\err{0.8} & 59.8\err{1.0} & 36.5\err{4.1} & 40.1\err{1.5} & 53.2 \\
Claude-4-sonnet & 73.6\err{2.6} & 70.1\err{0.7} & 63.4\err{1.7} & 45.3\err{2.8} & 34.6\err{1.6} & 57.4 \\
Claude-4-sonnet-thinking & 69.0\err{3.2} & 68.4\err{1.0} & 64.4\err{1.9} & 51.0\err{2.3} & 38.3\err{2.8} & 58.2 \\

\midrule
\multicolumn{7}{c}{\textit{Open LLMs (Prompting)}} \\
\midrule
Qwen2.5-14B-Instruct & 8.7\err{3.1} & 48.4\err{2.2} & 35.3\err{3.0} & 26.0\err{3.1} & 17.6\err{1.0} & 27.2 \\
Qwen2.5-32B-Instruct & 32.1\err{3.9} & 55.8\err{0.6} & 33.8\err{1.5} & 37.0\err{1.5} & 27.5\err{2.3} & 37.2 \\
Qwen2.5-72B-Instruct & 47.5\err{3.3} & 45.3\err{0.9} & 26.5\err{3.1} & 49.5\err{3.5} & 35.4\err{2.7} & 40.8 \\
\midrule
\multicolumn{7}{c}{\textit{Open LLMs (Agent Training)}} \\
\midrule
Hephaestus-8B-Base & 30.0 & 32.3 & 16.0 & 20.8 & 60.5$^{*}$ & 31.9 \\
Hephaestus-8B-IFT & 46.0 & 29.7 & 21.2 & 20.8 & 63.9$^{*}$ & 36.3 \\
AgentLM-7B & 84.0 & 30.6 & 18.1 & 17.4 & 63.6$^{*}$ & 42.7 \\
AgentLM-13B & 76.0 & 33.7 & 26.8 & 18.1 & 70.8$^{*}$ & 45.1 \\
AgentLM-70B & 86.0 & 37.7 & 47.0 & 21.5 & 64.9$^{*}$ & 51.4 \\
\midrule
\multicolumn{7}{@{}l}{\model} \\
\quad w/ Qwen2.5-3B-Instruct (RL) & 92.4\err{0.5} & 60.0\err{1.1} & 55.0\err{2.0} & 40.5\err{0.9} & 52.1\err{0.9} & 60.0 \\
\quad w/ Qwen2.5-7B-Instruct (RL) & 91.5$^{\dagger}$\err{0.9} & 63.7\err{0.5} & 57.8\err{2.3} & 40.8\err{1.2} & 56.1\err{0.6} & 62.0 \\
\quad w/ Qwen2.5-14B-Instruct (RL) & 91.5\err{0.9} & {72.2\err{0.9}} & 72.8\err{1.8} & 43.6\err{1.9} & 58.5\err{1.2} & {67.7} \\
\quad w/ Qwen2.5-32B-Instruct (RL) & \bfseries 94.5 \err{0.5}&  70.4 \err{0.5} & \bfseries 77.0 \err{1.2} & 51.7\err{1.8} & { 58.6 \err{0.9}} & \bfseries 70.4 \\
\quad w/ GLM-4-9B-0414 (SFT+RL)&  {93.3 \err{0.5}} & 66.9 \err{0.4} & \ulsi {75.7 \err{1.8}} & 33.2 \err{1.7} & 55.9 \err{1.9}& 65.0 \\
\midrule
\multicolumn{7}{c}{\textit{Single-Task Training (Qwen2.5-14B-Instruct)}} \\
\midrule
\model-ALFWorld & 89.7\err{1.6} & 49.7\err{1.6} & 22.3\err{3.1} & 33.7\err{3.1} & 15.9\err{0.5} & 42.3 \\
\model-DB & 0.2\err{0.5} & \bfseries 73.9\err{0.7} & 26.2\err{1.7} & 43.1\err{1.3} & 16.0\err{0.9} & 31.9 \\
\model-KG & 4.6\err{1.1} & 57.6\err{0.8} & 72.2\err{1.5} & 40.3\err{2.4} & 19.5\err{2.0} & 38.8 \\
\model-OS & 5.7\err{1.2} & 58.2\err{1.2} & 25.3\err{1.6} & 39.8\err{1.8} & 22.0\err{2.3} & 30.2 \\
\model-Webshop & 0.0\err{0.0} & 57.9\err{2.6} & 30.7\err{2.2} & 40.1\err{0.7} & \bfseries 60.3\err{1.3} & 37.8 \\
\midrule
Best-of-Single & 89.7\err{1.6} & \bfseries 73.9\err{0.7} & 72.2\err{1.5} & 43.1\err{1.3} & \bfseries 60.3\err{1.3} & \ulsi{67.8} \\
\bottomrule
\end{tabular*}
\begin{tablenotes}
\footnotesize
\item[$^\dagger$] We provide a one-shot demonstration for Qwen2.5-7B-Instruct in ALFWorld evaluation, as it fails to generate valid tool call format in the  environment.
\end{tablenotes}
\end{threeparttable}
\vspace{-5mm}
\end{table*}
}

\begin{table*}[t]
\caption{Multi-Task vs. Single-Task with Qwen2.5-14B-Instruct.}
\vspace{-2mm}
\label{tab:single-task}
\centering
\small
\footnotesize
\begin{threeparttable}
\renewcommand{\arraystretch}{1} %
\begin{tabular*}{\columnwidth}{@{}l @{\extracolsep{\fill}} S[table-format=2.1,detect-weight] S[table-format=2.1,detect-weight] S[table-format=2.1,detect-weight] S[table-format=2.1,detect-weight] S[table-format=2.1,detect-weight] S[table-format=2.1,detect-weight] @{}}
\toprule
\textbf{Model} & {\textbf{ALFWorld}} & {\textbf{DB}} & {\textbf{KG}} & {\textbf{OS}} & {\textbf{Webshop}} & {\textbf{AVG}} \\
\midrule
\model-ALFWorld & 89.7\err{1.6} & 49.7\err{1.6} & 22.3\err{3.1} & 33.7\err{3.1} & 15.9\err{0.5} & 42.3 \\
\model-DB & 0.2\err{0.5} &  73.9\err{0.7} & 26.2\err{1.7} & 43.1\err{1.3} & 16.0\err{0.9} & 31.9 \\
\model-KG & 4.6\err{1.1} & 57.6\err{0.8} & 72.2\err{1.5} & 40.3\err{2.4} & 19.5\err{2.0} & 38.8 \\
\model-OS & 5.7\err{1.2} & 58.2\err{1.2} & 25.3\err{1.6} & 39.8\err{1.8} & 22.0\err{2.3} & 30.2 \\
\model-Webshop & 0.0\err{0.0} & 57.9\err{2.6} & 30.7\err{2.2} & 40.1\err{0.7} &  60.3\err{1.3} & 37.8 \\
\midrule
Best of Five Models Above & 89.7\err{1.6} & 73.9\err{0.7} & 72.2\err{1.5} & 43.1\err{1.3} & 60.3\err{1.3} & \bfseries {67.8} \\
\midrule
\model (One Model) & 91.5\err{0.9} & 72.2\err{0.9} & 72.8\err{1.8} & 43.6\err{1.9} & 58.5\err{1.2} & \bfseries {67.7} \\
\bottomrule
\end{tabular*}

\end{threeparttable}
\vspace{-5mm}
\end{table*}

\textbf{Data.} 
We accommodate five agentic tasks (ALFWorld, DB, KG, OS, WebShop)~\citep{liu2024agentbench} to the \model infrastructure. 
The details of the dataset construction and unifying the function-call format are provided in Appendix \ref{sec:ds-detail}. 
To ensure that all tasks are sampled uniformly during training, we replicate smaller datasets such that each task appears approximately the same number of times as the largest task. 
Specifically, we sequentially cycle through multiple datasets, yielding one element from each in turn to produce interleaved output samples.

\textbf{Baselines.}
The closed-source API-based baselines include Claude-Sonnet~\citep{claude4}, GPT-5~\citep{gpt5}, and o-series models~\citep{o-series}. 
The general open models adopted include Qwen2.5-Instruct series (14B, 32B, and 72B)~\citep{qwen25}and DeepSeek-V3~\citep{liu2024deepseek} and R1~\citep{deepseekai2025deepseekr1incentivizingreasoningcapability}. 
We also compare against agent training methods on \agentbench, including Hephaestus~\citep{zhuang2025hephaestus} and AgentLM~\citep{zeng2024agenttuning}.

\subsection{Main Results}

We apply \model on open models, including Qwen2.5-Instruct series and GLM-4-9B-0414. 
Note that there is \textit{no warm-up supervised fine-tuning} before applying \model to all Qwen models. 
The main results are listed in Table~\ref{tab:main}. 

\textbf{SOTA Performance.}
Our \model framework achieves state-of-the-art performance across five tasks in \benchmark (see Appendix~\ref{sec:ds-detail}), establishing a new top average success rate of 70.4\%. Compared to the original Qwen2.5-Instruct models under prompting, \model yields substantial improvements, highlighting the effectiveness of reinforcement learning training. Notably, all \model-trained models, from 3B to 32B, consistently outperform strong baselines including leading models such as GPT-5, Claude-Sonnet-4 Thinking, and DeepSeek-R1.

\textbf{Multi-Task vs. Single-Task.}
Table~\ref{tab:single-task} shows that single-task RL agents excel only in their specific training environment but fail to generalize, yielding poor transfer across tasks. In contrast, our multi-task \model achieves nearly identical performance to the “best-of-five” single-task specialists while maintaining strong results on all tasks simultaneously. This highlights the effectiveness of multi-task training in acquiring generalizable skills without sacrificing peak performance.

\begin{table*}[t]
\centering
\small
\caption{Generalization Performance on BFCL-v3. 
}
\vspace{-2mm}
\label{tab:bfcl}
\setlength{\tabcolsep}{12pt} %
\resizebox{\columnwidth}{!}{
\begin{tabular}{
    @{}
    l
    c@{\hspace{1.2mm}}l
    c@{\hspace{1.2mm}}l
    c@{\hspace{1.2mm}}l
    c@{\hspace{1.2mm}}l
    @{}
}
\toprule
\multirow{2}{*}{\textbf{Model}}
& \multicolumn{4}{c}{\textbf{single-turn}}
& \multicolumn{2}{c}{\multirow{2}{*}{\textbf{multi-turn}}}
& \multicolumn{2}{c}{\multirow{2}{*}{\textbf{overall}}} \\
\cmidrule(lr){2-5}
& \multicolumn{2}{c}{\textbf{nonlive}}
& \multicolumn{2}{c}{\textbf{live}}
& \multicolumn{2}{c}{} & \multicolumn{2}{c}{} \\
\midrule
Qwen2.5-32B-Instruct
 & 86.0\err{0.2} &
 & 77.4\err{0.1} &
 & 16.2\err{0.6} &
 & 59.9 &
\\
\model w/ Qwen2.5-Instruct-32B  
 & 85.8\err{0.2} & \textcolor{darkred}{\tiny$\downarrow$0.2}
 & 79.3\err{0.2} & \textcolor{black}{\tiny$\uparrow$1.9}
 & 19.2\err{0.8} & \textcolor{black}{\tiny$\uparrow$3.0}
 & 61.4 & \textcolor{black}{\tiny$\uparrow$1.5}
\\
\bottomrule
\end{tabular}
}
\vspace{-5mm}
\end{table*}

\textbf{Generalization on BFCL-v3.}
To examine generalization, we evaluate the \model model (trained on ALFWorld, DB, KG, OS, and Webshop) on the BFCL-v3 benchmark~\citep{patil2025bfcl}.  
As shown in Table~\ref{tab:bfcl}, \model demonstrates clear improvements on multi-turn tasks and modest gains on single-turn tasks. These results suggest that our approach can enhance the generalizability of function calling, providing a step toward more broadly capable agentic LLMs.

\hide{

We describe our experiment setup and results in this section.

\vpara{Dataset Construction.} We updated the original \agentbench\ to accommodate the new infrastructure and to unify the function-call format. This modified version, which we refer to as \benchmark, additionally incorporates a synthetic training dataset. Further details of the modifications and dataset construction are provided in the Appendix \ref{sec:ds-detail}.

\vpara{Balanced Sampling.} To ensure that all tasks are sampled uniformly during training, we replicate smaller datasets such that each task appears approximately the same number of times as the largest task. Specifically, we sequentially cycle through multiple datasets, yielding one element from each in turn to produce interleaved output samples.

\vpara{Unified Reward Scheme.} We normalize all task rewards to the range $[0, 1]$ for consistency. For tasks without intrinsic reward signals, we assign a reward of 1 for correct responses and 0 otherwise. In addition, we leverage termination signals provided by \benchmark and penalize abnormal terminations with a reward of $-0.2$ to encourage proper episode completion.

\subsection{Main Results}

We performed our train-from-scratch experiments primarily on the Qwen2.5 series of models, with the main results presented in Table~\ref{tab:main}. Our method, termed \model, achieves state-of-the-art overall performance, demonstrating the effectiveness and efficiency of our designed training system.

\textbf{Baselines.}
To provide a comprehensive evaluation, we established a robust set of baselines. This includes powerful, closed-source API-based models such as Claude-4-sonnet, GPT-5, and DeepSeek-R1, which serve as high-standard baselines for agent performance. We also include the Qwen2.5-Instruct series (14B, 32B, and 72B) as essential comparison points, thereby enabling a clear assessment of the performance gains from our RL training. Furthermore, we compare against previously established agent training methods on \agentbench, such as Hephaestus-8B-IFT and AgentLM \citep{zeng2024agenttuning}, to contextualize our contributions within the existing literature.

\textbf{Multi-task Training.}
Our primary experiments involve multi-task RL across all five \benchmark environments. The results show that \model significantly outperforms all baselines in terms of average score. Notably, our \model-Qwen2.5-32B-Instruct model achieves an average score of 70.4, representing a substantial improvement of 12.2 points over the strongest API baseline, Claude-4-sonnet-thinking (58.2). Our model not only excels on average but also achieves the best performance on four out of the five individual tasks: ALFWorld (94.5), DB (70.4), KG (77.0), and Webshop (58.6). This demonstrates the strong multi-task capability of our method.

\textbf{Single-task Training.}
To further analyze the benefits of multi-task learning, we conducted single-task training experiments, where an agent was trained exclusively on one environment with the same method. The results indicate that single-task RL can achieve very strong performance on its specific target task. However, these specialist agents exhibit poor generalization to other tasks, as evidenced by the low transfer scores. The results show that our multi-task training experiments can match the performance of single-task experiments at the same model scale, demonstrating that our method successfully acquires multi-task ability without performance drop.

}

\begin{table}[htbp]
\caption{Ablation on cross-policy sampling and task advantage normalization. }
\label{tab:ablation_study}
\centering
\small
\footnotesize
\setlength{\tabcolsep}{12pt} %
\renewcommand\arraystretch{.95}
\resizebox{\columnwidth}{!}{
\begin{tabular}{@{}lcccccc@{}}
\toprule
Method & AF & DB & KG & OS & WS & AVG \\
\midrule
\model -14B &  93.1\err{0.5} & 64.0\err{0.5} & 67.7\err{2.0} & 45.1\err{2.0} & 55.0\err{0.7} & 65.0 \\
{- cross sampling} & 91.9\err{1.2} & 61.6\err{1.0} & 55.7\err{1.4} & 39.7\err{2.3} & 54.5\err{1.3} & 60.7 \\
{- task adv. norm} & 91.1\err{0.9} & 62.6\err{0.7} & 54.7\err{1.6} & 38.0\err{2.0} & 50.6\err{1.7} & 59.4  \\
\bottomrule
\end{tabular}
}
\end{table}

\begin{figure}[htbp]
    \centering
    \begin{subfigure}[t]{0.32\linewidth}
        \centering
        \includegraphics[width=\linewidth,keepaspectratio]{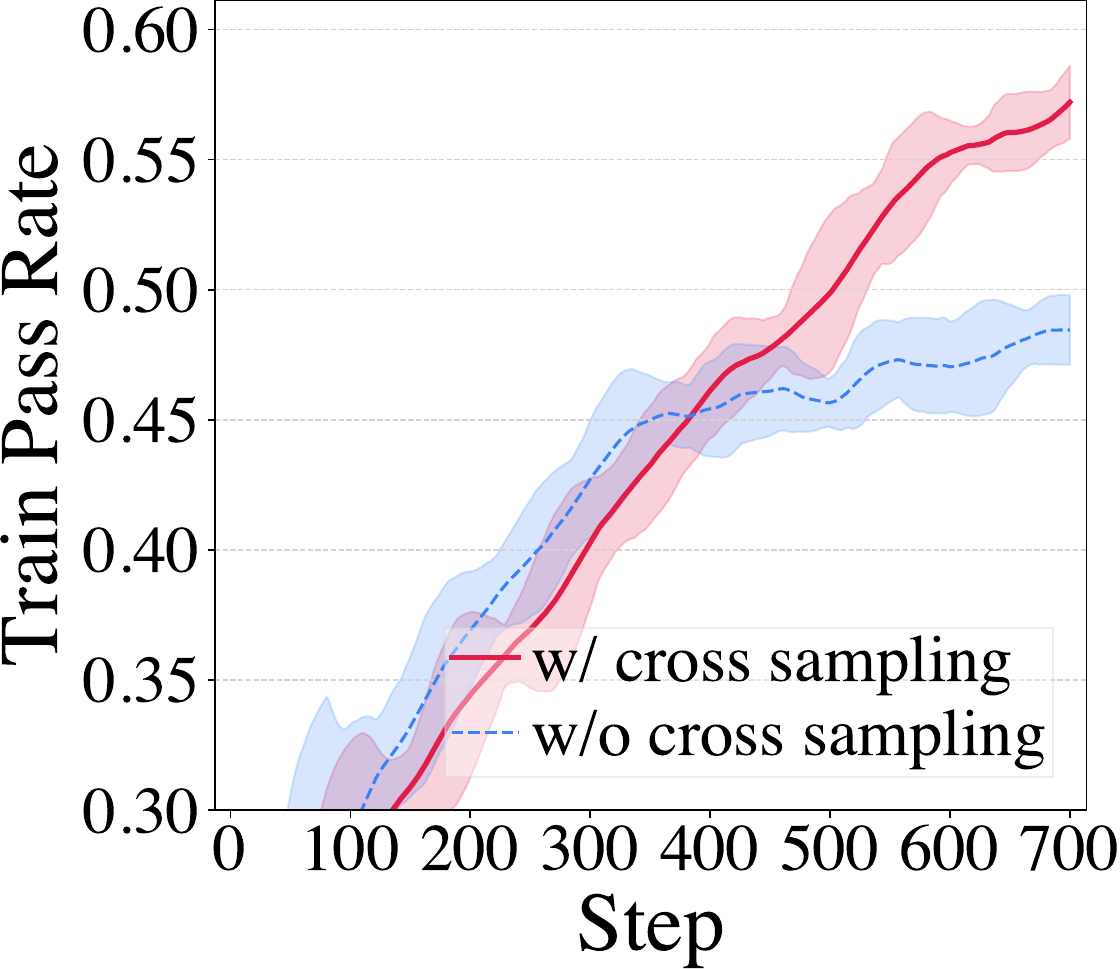}
        \caption{Cross-Policy Sampling in KG.}
        \label{fig:kg_ablation}
    \end{subfigure}\hfill
    \begin{subfigure}[t]{0.32\linewidth}
        \centering
        \includegraphics[width=\linewidth,keepaspectratio]{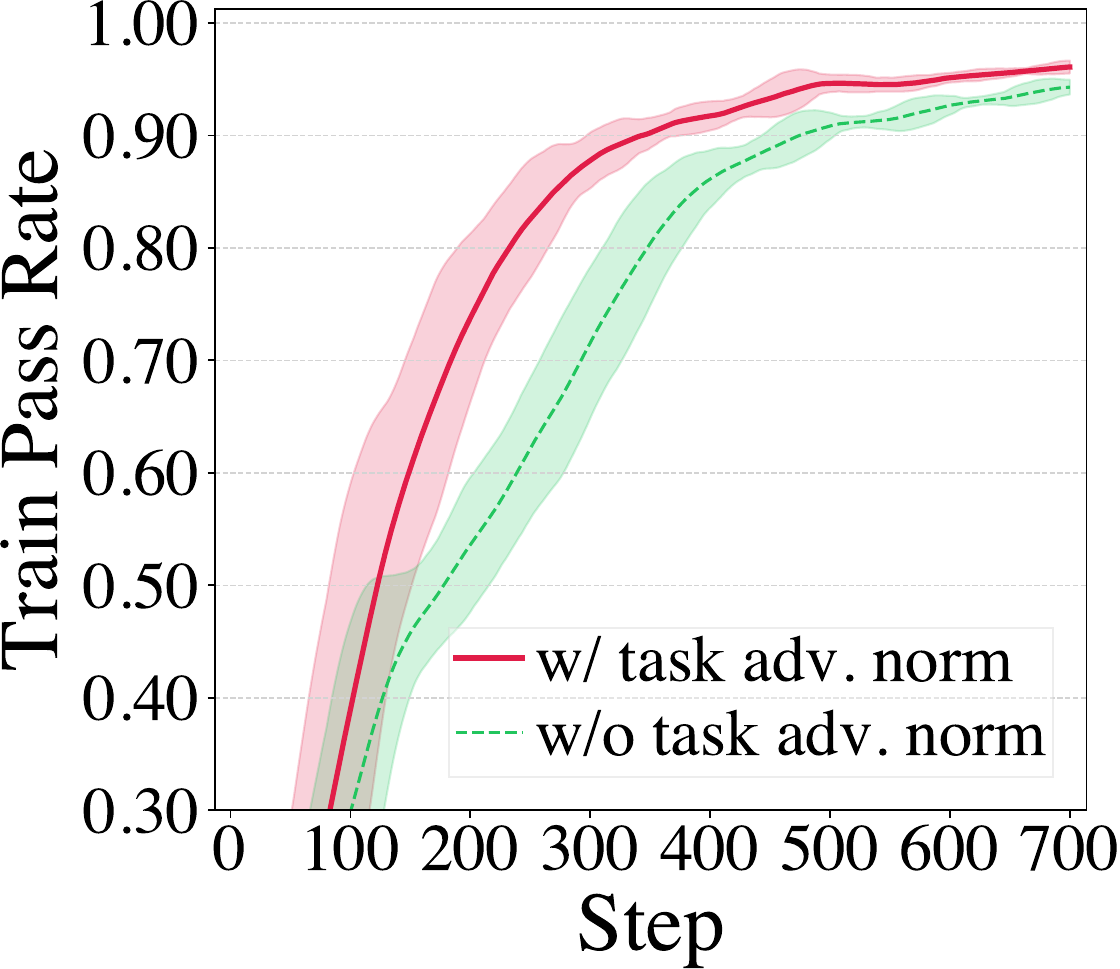}
        \caption{Task Adv. Norm. in ALFWorld.}
        \label{fig:alf_ablation}
    \end{subfigure}\hfill
    \begin{subfigure}[t]{0.32\linewidth}
        \centering
        \includegraphics[width=\linewidth,keepaspectratio]{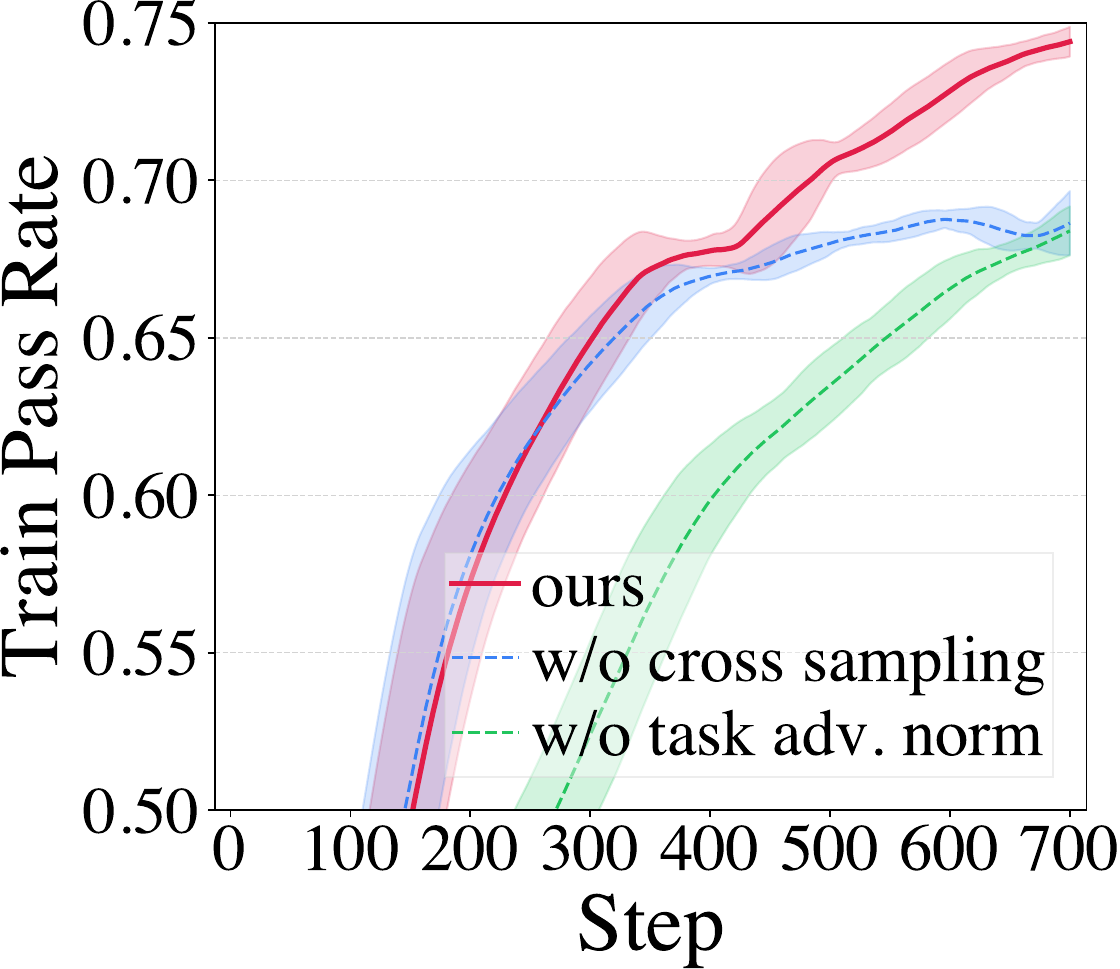}
        \caption{Average over 5 environments.}
        \label{fig:all_ablation}
    \end{subfigure}
    \caption{Ablation studies. 
    (c): The combined effect of Cross-Policy Sampling and Task Advantage Normalization, averaged over five environments.}
    \label{fig:ablation_studies_combined}
    \vspace{-5mm}
\end{figure}

\subsection{Ablation Study}

\textbf{Cross-Policy Sampling.}
Table \ref{tab:ablation_study} suggests \model trained without cross-policy sampling performs worse. 
This phenomenon is especially obvious in some tasks/environments. 
We demonstrate the pass rate on KG during training in Figure \ref{fig:kg_ablation} as an example; the model's capability reaches the top earlier than the model trained with  cross-policy sampling. 
These results demonstrate that cross-policy sampling is able to explore more possible states, especially in more open-ended environments during training, thus expanding the border of the model's capability.

\textbf{Task Advantage Normalization.}
Table~\ref{tab:ablation_study} suggests that removing task advantage normalization leads to clear performance drops. 
Also, as shown in Figure~\ref{fig:alf_ablation}, the training efficacy is severely reduced and demonstrates fluctuations on some tasks. 
When removing the task advantage normalization, the model tends to learn different tasks at different rates instead of learning jointly.
These results indicate that normalizing the advantage for each task effectively stabilizes multi-task training and reduces negative interference, resulting in more robust and consistent learning across tasks.

\subsection{Verifying the Effect of the Cross-Policy Sampling Strategy}
\label{sec:cross-exp}

\begin{figure}[htbp]
    \centering
    \begin{subfigure}[t]{0.48\linewidth}
        \centering
        \includegraphics[width=\linewidth]{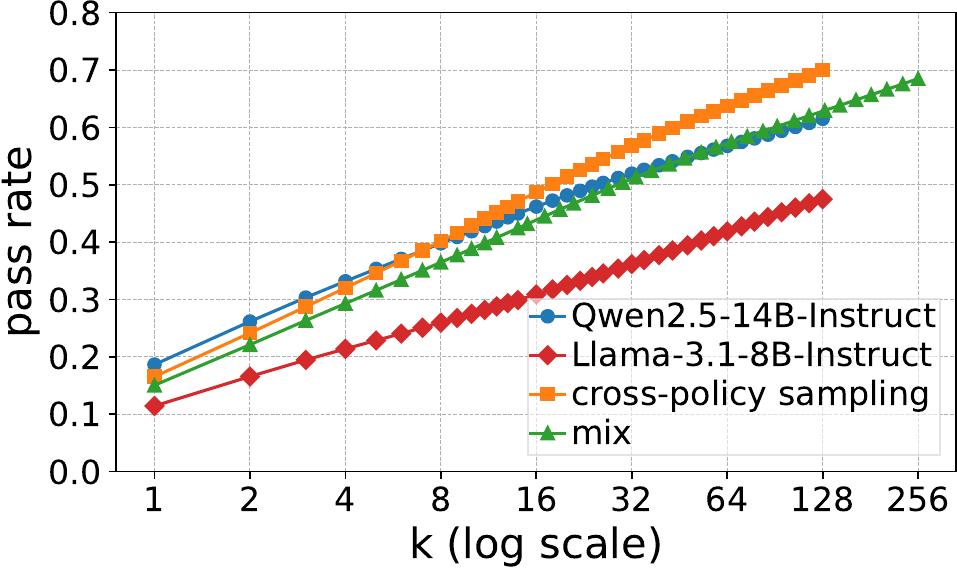}
        \caption{
        Cross-policy sampling on Webshop.
        The \textit{mix} strategy combines data from both models, so its maximum $K$ is twice that of the other strategies. }
        \label{fig:qwen_llama_passak}
        \vspace{-1mm}
    \end{subfigure}
    \quad
    \begin{subfigure}[t]{0.48\linewidth}
        \centering
        \includegraphics[width=\linewidth]{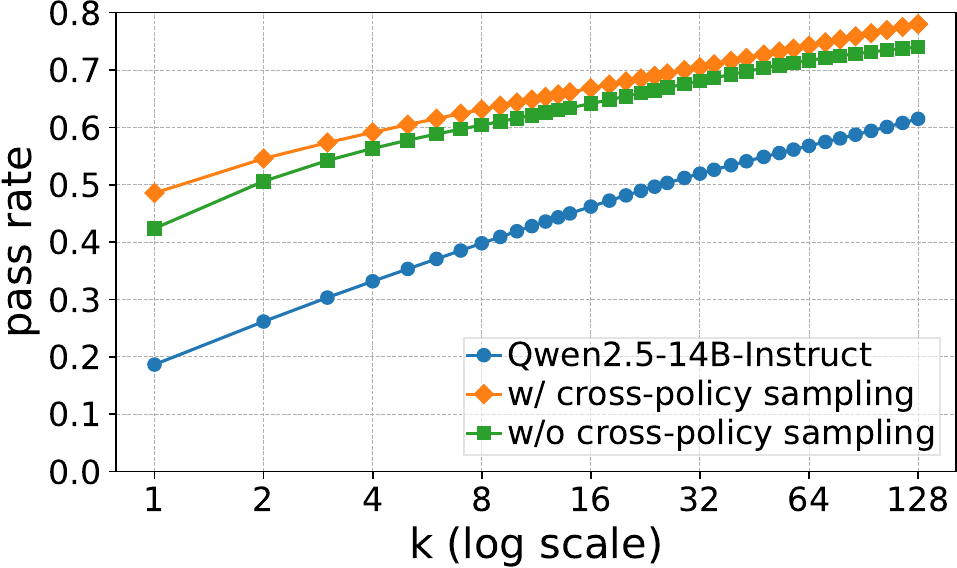}
        \caption{Results from preliminary experiments on the WebShop environment. Note that settings are not completely the same as those in the main experiments.}
        \label{fig:cs_train_compare}
        \vspace{-1mm}
    \end{subfigure}
    \caption{Effects of   cross-policy sampling in inference (a) and training (b)  on Webshop. %
    }
\end{figure}

\textbf{Applying Cross-Policy sampling in Inference. }
The proposed cross-policy sampling strategy samples actions from a pool of models (as depicted in Figure~\ref{fig:rollouts}). To verify that the cross-policy sampling strategy effectively promotes model exploration, we first directly applied our method to inference. We conducted experiments using the Qwen~\citep{qwen25} and Llama~\citep{grattafiori2024llama3herdmodels} models in the WebShop~\citep{webshop} environment. 
As shown in Figure \ref{fig:qwen_llama_passak}, we observe that in low-$k$ regimes, the performance of the cross-policy sampling strategy is slightly lower than the best single model strategy. However, as $k$ increases, a surprising trend emerges: the cross-policy sampling strategy eventually surpasses both individual models in \texttt{pass@k} metrics.
The performance of the cross-policy sampling strategy also surpasses mixing two models' trajectories, demonstrating that the strategy has effectively explored something outside both models' capability boundaries. This provides strong evidence for our theoretical analysis.

\textbf{Applying Cross-Policy sampling in RL. }
To further verify the effectiveness of the cross-policy sampling strategy during RL training, we conduct a training experiment on the Webshop task. %
As shown in Figure \ref{fig:cs_train_compare}, both trained models demonstrated a significant improvement in \texttt{pass@1} rate compared to the untrained base model. But the model trained with the cross-policy sampling strategy demonstrates a consistent advantage as $k$ increases. 
This suggests that the strategy successfully preserves the model's diversity while improving its overall ability.

\section{Related Work}
\textbf{Reinforcement Learning AI Agents.}
RL algorithms like PPO~\citep{schulman2017proximal} and GRPO~\citep{shao2024deepseekmathpushinglimitsmathematical} have been widely adopted in LLM agent training. Deepseek-R1~\citep{deepseekai2025deepseekr1incentivizingreasoningcapability} demonstrates RL’s ability to incentivize reasoning in LLMs through reward-driven fine-tuning.
Recent works~\citep{qian2025toolrl,feng2025group,lu2025arpo,wen2025reinforcementlearningverifiablerewards} further develop RL techniques.
GUI agents also benefit from RL-driven optimization~\citep{xu2024androidlabtrainingsystematicbenchmarking,qi2024webrl,liu2024autoglmautonomousfoundationagents,qin2025uitarspioneeringautomatedgui,chen2025reinforcementlearninglonghorizoninteractive}.
For long-horizon tasks, \citet{chen2025reinforcementlearninglonghorizoninteractive} shows RL’s efficacy in balancing exploration and tool usage. DeepResearcher further scales real-world research by training agents to iteratively refine hypotheses via RL~\citep{zheng2025deepresearcherscalingdeepresearch}.
Despite these advancements, most current approaches fall short in studying the exploration aspect of RL training and the multi-task setting. 
In this work, we propose the cross-policy sampling strategy and task advantage normalization, addressing a critical gap in existing methods.

\textbf{Reinforcement Learning Infrastructure.}
Several frameworks~\citep{sheng2024hybridflow,hu2024openrlhf,fu2025areal} have been developed for RL training. 
These frameworks usually adopt modern training~\citep{shoeybi2019megatron,zhao2023pytorch} and rollout~\citep{kwon2023efficient,zheng2024sglang} engines to boost efficiency.
However, unlike math or coding tasks, agent scenarios involve multi-turn interactions with environments. 
There have been works~\citep{liu2024agentbench,ma2024agentboardanalyticalevaluationboard} to provide standardized benchmarks for evaluating multi-turn interactions and addressing reproducibility gaps. 
Platforms such as E2B~\citep{e2b2025} and OpenHands~\citep{wang2025openhandsopenplatformai} provide secure sandbox environments and modular interfaces for code execution, browser automation, and generalist agent development.
While these environments provide strong support for agent evaluation, existing RL frameworks lack built-in support for multi-turn interactions and agent-specific training optimizations.

\section{Conclusion}

We propose \model, a system for training LLM agents with RL across diverse tasks and environments. 
Through asynchronous rollout–training pipelines, scalable environment deployment, and algorithmic advances including cross-policy sampling and task advantage normalization, \model enables more efficient and stable training. Experiments demonstrate competitive results across diverse agentic benchmarks, with encouraging signs of generalization to unseen tasks.

\bibliographystyle{plainnat}
\bibliography{ref}

\clearpage

\appendix
\section{Background of Reinforcement Learning in Large Language Models}
Reinforcement Learning (RL) has significantly enhanced the capabilities of Large Language Models (LLMs) by optimizing their decision-making through reward-driven training. The fundamental RL objective is expressed as:
\begin{equation}
\mathcal{J}(\theta) = \mathbb{E}_{s \sim \mathcal{D}, a \sim \pi_{\theta}(s)} [R(s, a)],
\end{equation}
where \(\pi_{\theta}\) denotes the policy, \(s\) represents the input context, \(a\) is the generated output, and \(R(s, a)\) assesses the output quality via a reward function.

\textbf{A key method, Proximal Policy Optimization (PPO)}\citep{schulman2017proximal}, ensures training stability using a clipped probability ratio, defined as:
\begin{equation}
\rho_t(\theta) = \frac{\pi_{\theta}(a_t|s_t)}{\pi_{\text{old}}(a_t|s_t)},
\end{equation}
with the objective function:
\begin{equation}
\mathcal{J}_{\text{PPO}}(\theta) = \mathbb{E}_t [\min(\rho_t(\theta) \hat{A}_t, \text{clip}(\rho_t, 1 - \epsilon, 1 + \epsilon) \hat{A}_t) - \beta D_{\text{KL}}],
\end{equation}
where \(\hat{A}_t\) is the advantage estimate, and clipping limits policy updates.

For improved advantage estimation, Generalized Advantage Estimation (GAE)\citep{schulman2018highdimensionalcontinuouscontrolusing} is utilized, computed as:
\begin{equation}
\hat{A}_t^{\text{GAE}}(\gamma, \lambda) = \sum_{l=0}^{\infty} (\gamma \lambda)^l \delta_{t+l},
\end{equation}
where \(\delta_t = r_t + \gamma V(s_{t+1}) - V(s_t)\) is the temporal difference error, and \(\gamma\) and \(\lambda\) adjust the bias-variance tradeoff.

\textbf{Another approach, Group Relative Policy Optimization (GRPO)}\citep{shao2024deepseekmathpushinglimitsmathematical}, optimizes over groups of outputs with the objective:
\begin{equation}
\mathcal{J}_{\text{GRPO}}(\theta) = \mathbb{E}_{o_i \sim \pi_{\text{group}}(\theta)} [J_{\text{group}}(\theta)],
\end{equation}
where the group objective is:
\begin{equation}
\mathcal{J}_{\text{group}}(\theta) = \frac{1}{G} \sum_{i=1}^{G} \min(\rho_i \hat{A}_i, \hat{\rho}_i) - \beta D_{\text{KL}},
\end{equation}
and the advantage \(\hat{A}_i\) is a normalized reward:
\begin{equation}
\hat{A}_i = \frac{r_i - \mu_r}{\sigma_r},
\end{equation}
with \(\mu_r\) and \(\sigma_r\) as the mean and standard deviation of rewards, fostering adaptive LLM behaviors.

\textbf{Finally, Decoupled Clip and Dynamic sampling Policy Optimization (DAPO)}\citep{yu2025dapoopensourcellmreinforcement} was proposed to address issues specific to long-CoT reinforcement learning, such as entropy collapse and training instability. The algorithm modifies the GRPO objective by introducing several key techniques, including a decoupled clipping mechanism and a dynamic sampling strategy.

The DAPO objective function is formulated as:
\begin{multline}
\mathcal{J}_{DAPO}(\theta) = \mathbb{E}_{(q,a)\sim\mathcal{D},\{o_{i}\}_{i=1}^{G}\sim\pi_{\theta_{old}}(\cdot|q)} \biggl[ \frac{1}{\sum_{i=1}^{G}|o_{i}|}\sum_{i=1}^{G}\sum_{t=1}^{|o_{i}|} \\
\times \min \Bigl( r_{i,t}(\theta)\hat{A}_{i,t}, \text{clip}(r_{i,t}(\theta),1-\epsilon_{low},1+\epsilon_{high})\hat{A}_{i,t} \Bigr) \biggr]
\end{multline}
subject to the constraint:
\begin{equation}
0<|\{o_{i}|\text{is\_equivalent}(a,o_{i})\}|<G,
\end{equation}

where the advantage $\hat{A}_{i,t}$ is calculated similarly to GRPO. The primary innovations are the \textbf{decoupled clipping bounds}, $\epsilon_{low}$ and $\epsilon_{high}$, which allow for greater exploration to prevent entropy collapse, and the \textbf{dynamic sampling constraint}, which filters out batches where all responses are either correct or incorrect to ensure a non-zero advantage and stable gradients. The loss is also normalized at the \textbf{token level} ($\frac{1}{\sum|o_i|}$) to properly weight responses of varying lengths.

\section{Preliminaries}
\label{sec:pre}

\subsection{Problem Definition}

\paragraph{Agentic Task.}
We define an \emph{agentic task} $\mathcal{T}_i$ as a Markov Decision Process (MDP):
\begin{equation}
\mathcal{T}_i = \big(\mathcal{S}_i^{\mathrm{env}}, \mathcal{A}_i, P_i, r_i, \rho_i \big),
\end{equation}
where $\mathcal{S}_i^{\mathrm{env}}$ is the environment state space, $\mathcal{A}_i$ is the action space,
$P_i(s'|s,a)$ denotes transition dynamics, $r_i(s,a)$ is the reward function,
and $\rho_i(s_0)$ is the initial state distribution.

\paragraph{LLM-based Policy and Composite State.}
When the policy $\pi_\theta$ is implemented by an LLM,
the state at decision step $t$ is the \emph{composite state} 
$s_t = (s_t^{\mathrm{env}}, s_t^{\mathrm{ctx}})$, 
where $s_t^{\mathrm{env}}$ is the environment state and 
$s_t^{\mathrm{ctx}} \in \mathcal{V}^*$ is a tokenized context representing the 
trajectory prefix up to step $t$.

In LLM-based settings, a high-level action $a_t$ is a complete sequence of 
$L_t$ tokens:
\begin{equation}
    a_t = (y_{t,1}, y_{t,2}, \dots, y_{t,L_t}), \quad y_{t,k} \in \mathcal{V}.
\end{equation}
The underlying LLM defines a token-level probability distribution
$P_\theta(y_{t,k} \mid s_t^{\mathrm{ctx}}, y_{t,<k})$ for each token, 
and the policy probability of producing $a_t$ from $s_t$ factorizes as:
\begin{equation}
    \pi_\theta(a_t \mid s_t) = 
    \prod_{k=1}^{L_t} P_\theta\left( y_{t,k} \mid s_t^{\mathrm{ctx}}, y_{t,<k} \right).
\end{equation}
This factorization allows us to define token-level log-probabilities and, 
consequently, token-level policy gradients and advantage estimates.

\paragraph{Trajectory Definition.}
A trajectory in task $\mathcal{T}_i$ is defined as
\begin{equation}
\label{eq:traj}
\tau = \big( s^{(0)}, a^{(0)}, r^{(1)}, s^{(1)}, a^{(1)}, r^{(2)}, \dots, s^{(T-1)}, a^{(T-1)}, r^{(T)}, s^{(T)} \big),
\end{equation}
where each $s^{(t)} = (s_t^{\mathrm{env}}, s_t^{\mathrm{ctx}})$ is a composite state as above.
The reward $r^{(t+1)} = r_i(s_t^{\mathrm{env}}, a^{(t)})$ 
is assigned after $a^{(t)}$ is applied in $s_t^{\mathrm{env}}$.
Different from standard MDP trajectories, this formulation explicitly embeds a 
context component in each state.

\paragraph{Multi-Task Setting.}
We study a collection of $N_{\mathrm{task}}$ tasks:
\begin{equation}
\mathcal{T} = \{\mathcal{T}_1, \ldots, \mathcal{T}_{N_{\mathrm{task}}}\}.
\end{equation}
For each $\mathcal{T}_i$ there are $M_i$ samples:
\begin{equation}
\mathcal{D}_i = \{ x_{i,1}, \dots, x_{i,M_i} \}.
\end{equation}
Executing sample $x_{i,j}$ produces a \emph{group} of $K_{i,j}$ trajectories:
\begin{equation}
G_{i,j} = \{ \tau_{i,j,1}, \dots, \tau_{i,j,K_{i,j}} \},
\end{equation}
which, as we will discuss in RLVR, are used in GRPO to compute group-based advantage estimates.

\subsection{Reinforcement Learning with Verifiable Rewards (RLVR)}
\label{sec:rlvr}

Reinforcement Learning with Verifiable Rewards (RLVR)~\cite{deepseekai2025deepseekr1incentivizingreasoningcapability} refers to scenarios in which the reward signal associated with a trajectory can be computed in a deterministic and objective manner based on the observed interaction data. %
In practice, RLVR is commonly optimized using 
Proximal Policy Optimization (PPO)~\cite{schulman2017proximal} or its extensions such as 
Group Relative Policy Optimization (GRPO)~\citep{shao2024deepseekmathpushinglimitsmathematical}. %

\paragraph{PPO Objective.}
Given a batch of trajectories, PPO maximizes the clipped surrogate objective:
\begin{equation}
\mathcal{L}_{\mathrm{PPO}}(\theta) =
\mathbb{E}_{t} \left[
\min\left(
r_t(\theta) \hat{A}_t,\ 
\mathrm{clip}(r_t(\theta), 1 - \epsilon, 1 + \epsilon)\hat{A}_t
\right)
\right],
\end{equation}
where
$r_t(\theta) = \frac{\pi_\theta(a_t \mid s_t)}{\pi_{\theta_{\mathrm{old}}}(a_t \mid s_t)}$
is the probability ratio and $\hat{A}_t$ is the advantage estimate.

\paragraph{GRPO Objective.}
Under GRPO, each group $G_{i,j}$ contains $K_{i,j}$ trajectories compared within the group,
yielding group-relative advantage estimates $\hat{A}_{i,j,g}$.  
The objective is:
\begin{equation}
\mathcal{L}_{\mathrm{GRPO}}(\theta) =
\mathbb{E}_{i,j} \left[
\frac{1}{K_{i,j}} \sum_{g=1}^{K_{i,j}}
\min\left(
\rho_{i,j,g}(\theta) \,\hat{A}_{i,j,g},\ 
\mathrm{clip}(\rho_{i,j,g}(\theta), 1 - \epsilon, 1 + \epsilon)\,\hat{A}_{i,j,g}
\right)
\right],
\end{equation}
where
$
\rho_{i,j,g}(\theta) =
\frac{\pi_\theta(a_{i,j,g} \mid s_{i,j,g})}
{\pi_{\theta_{\mathrm{old}}}(a_{i,j,g} \mid s_{i,j,g})}$
and 
$\hat{A}_{i,j,g} = \frac{\hat{R}_{i,j,g} - \text{mean}(\hat{R}_{i,j})}{\text{std}(\hat{R}_{i,j})}$
is a group-relative advantage estimate, computed as the difference between 
the empirical return $\hat{R}_{i,j,g}$ of trajectory $\tau_{i,j,g}$.

\section{Formal Description and Intuition of the Cross-Policy Sampling Strategy}
\label{sec:formal-cs}
Formally speaking, compared to Equation~\ref{eq:traj}, 
the trajectory obtained by cross sampling takes the form of:
\begin{equation}
    \tau^{c} = \big(s^{(0)}, a^{c,(0)}, r^{(1)}, s^{(1)}, a^{c,(1)}, r^{(2)}, \ldots, s^{(T)}\big)
    \quad \text{where} \quad a^{c,(t)} \sim \mathrm{random}(\mathcal{M})(\cdot \mid s^{(t)}),
    \label{eq:cross-traj}
\end{equation}
where each $s^{(t)} = (s_t^{\mathrm{env}}, s_t^{\mathrm{ctx}})$ is the composite state defined earlier,
and $\mathcal{M}$ is the set of models.

Intuitively speaking, the language state $s^\mathrm{ctx}$ stochastically maps to environmental states $s^{\mathrm{env}}$ through a grounding function $\Gamma: \mathcal{L}_{\text{valid}} \to \Delta(\mathcal{S}^{\mathrm{env}})$, where $\mathcal{L}_{\text{valid}} \subset \mathcal{L}$ is the space of linguistically coherent token sequences. For a success set $\mathcal{G} \subset \mathcal{S}^{\mathrm{env}}$, define its language preimage $\mathcal{L}_{\mathcal{G}} = \Gamma^{-1}(\mathcal{G}) \cap \mathcal{L}_{\text{valid}}$ -- the set of valid language states that can reach $\mathcal{G}$. 

Cross sampling expands coverage of $\mathcal{L}_{\mathcal{G}}$ while preserving validity: 
\[
\text{supp}(\tau^c) \cap \mathcal{L}_{\mathcal{G}} \supsetneqq \bigcup_m \left( \text{supp}(\tau^m) \cap \mathcal{L}_{\mathcal{G}} \right)
\]
where $\text{supp}(\tau)$ denotes language states visited in the trajectory $\tau$. This increases the probability $P(s^\mathrm{env} \in \mathcal{G})$ by exploring more paths in $\mathcal{L}_{\mathcal{G}}$, without deviating into $\mathcal{L} \setminus \mathcal{L}_{\text{valid}}$.

\section{Dataset Details}
\label{sec:ds-detail}
\subsection{Extending \agentbench}

While the overall framework is decoupled from benchmarks, we perform training on a refined version of \agentbench, or what we call \benchmark. Specifically, we made several modifications:

\subsection{Synthesizing Training Set}
\label{subsec:trainset}

To address the scarcity of training data in the original \agentbench framework, we aim to construct a large-scale and diverse dataset suitable for reinforcement learning across various agent environments. To this end, we adopted a multifaceted data collection strategy tailored to the unique characteristics of each environment:

\vpara{Direct Adoption of Existing Datasets.} For environments like \texttt{AlfWorld} and \texttt{WebShop}, which are accompanied by rich, pre-existing training sets, we directly incorporated these official datasets. This approach ensures consistency with the original benchmarks and leverages well-established data sources.

\vpara{Synthetic Data Generation via Self-Instruct.} For tasks in more complex environments such as \texttt{OS}, \texttt{KnowledgeGraph}, and \texttt{DB}, where training data is not readily available, we employed the Self-Instruct methodology \citep{wang2022self}. We used high-performance APIs (o3 and claude4-sonnet) to efficiently sample and filter a large volume of high-quality training instances.

\vpara{Augmentation with External High-Quality Datasets.} To further enrich the diversity and complexity of our training data, we integrated external, high-quality datasets. Notably, for the \texttt{DB} environment, we augmented our dataset with the training samples provided by the BIRD benchmark \citep{li2023llmservedatabaseinterface}, a comprehensive text-to-SQL dataset.

\subsection{Modifications to \agentbench Environment}

To enhance the flexibility and compatibility of \agentbench, we transformed its five environments into a Function-Call Based framework. We analyzed the distinct action types required by each environment and categorized them accordingly. For each environment, we extracted specific tools following the OpenAI Function Call Format. For instance, in the Knowledge Graph (KG) environment, we identified and implemented seven tools, including \texttt{get\_relations}, \texttt{get\_neighbors}, and \texttt{count}, among others. Additionally, we modified the interaction logic of each environment to support external requests in the Function Call format, ensuring seamless integration with external systems.

Outlining the refactoring of Controller and Worker interfaces
We restructured the interface protocols for the Controller and Worker components in \agentbench to standardize task management and interaction. The \texttt{start\_sample} interface was introduced to initiate a task, while multi-turn interactions were facilitated through the \texttt{interact} interface. To improve Controller oversight, we implemented additional interfaces, such as \texttt{list\_sessions} and \texttt{list\_workers}, enabling efficient monitoring of internal worker and session states within the container.

\section{Detailed Experimental Settings}
\subsection{Environments and Tasks}
We select five representative multi-turn interaction tasks from the AgentBench dataset~\cite{liu2024agentbench}, a comprehensive and evolving benchmark designed to evaluate the reasoning and decision-making capabilities of large language models. These tasks, encompassing operating system interactions, database management, knowledge graph navigation, text-based adventure games, and web shopping scenarios, are chosen for their diverse challenges and ability to assess critical skills such as long-sequence comprehension, contextual tracking, and environmental interaction. The tasks are supported by standardized evaluation protocols and open-source code environments, facilitating robust experimental implementation and framework refinement.

\textbf{Unified Reward.} 
We normalize all task rewards to the range $[0, 1]$ for consistency. For tasks without intrinsic reward signals, we assign a reward of 1 for correct responses and 0 otherwise. 
In addition, we leverage termination signals and penalize abnormal terminations with a reward of $-0.2$ to encourage proper episode completion.

\begin{itemize}[leftmargin=*,itemsep=0pt,parsep=0.2em,topsep=0.2em,partopsep=0.0em]
    \item \textbf{Operating System (OS) Task:} This environment assesses an agent's ability to interact with a real Ubuntu Docker-based operating system through Bash command-line inputs. Agents are tasked with interpreting natural language instructions and translating them into precise Shell commands to achieve specific objectives, such as file manipulation or directory navigation in an unfamiliar environment. The task demands high accuracy in command generation, error handling, and result interpretation (e.g., standard output and error streams), given the vast action space and the need for adaptive decision-making.

    \item \textbf{Database (DB) Task:} In this scenario, agents act as database analysts, interacting with a real database via SQL queries to address natural language questions or perform data modifications (e.g., INSERT, UPDATE). The task evaluates the agent's proficiency in converting natural language to SQL (Text-to-SQL), understanding database schemas (table structures, column names, data types), and managing complex queries (e.g., multi-table joins, nested queries, aggregation functions). Multi-turn interactions require agents to adjust strategies based on query results or error feedback.

    \item \textbf{Knowledge Graph (KG) Task:} For the KG environment, API results are obtained with one-shot testing to ensure the model can correctly invoke tool calls, while our trained models are trained and evaluated without one-shot assistance. This task challenges agents to perform multi-step reasoning and information retrieval within a large knowledge graph (e.g., Freebase) to answer complex queries. With only partial observability due to the graph's scale, agents must use structured query operations (e.g., retrieving entity relationships or finding intersecting entity sets via callable tools) to explore and connect information fragments. It emphasizes long-term planning, information integration, and effective decision-making under incomplete information.

    \item \textbf{Text Adventure Game (Text Game / House-Holding, HH - Represented by ALFWorld):} Agents operate in a text-described virtual household environment, executing action sequences to meet high-level goals (e.g., “clean a soapbar and place it on the workbench”). Actions include navigating (e.g., “go to cabinet 1”), interacting with objects (e.g., “take soapbar 1 from sinkbasin 1”), and adjusting plans based on feedback (e.g., “The cabinet 2 is closed”). ALFWorld~\citep{shridhar2020alfworld} highlights the need for commonsense reasoning, goal decomposition, and dynamic planning in response to environmental states.

    \item \textbf{Web Shopping (WS - Represented by WebShop):} This task simulates an e-commerce experience where agents search for products based on specific criteria (e.g., brand, price) by interacting with a simulated website. Actions include keyword searches, link clicks, attribute filtering, and adding items to a cart. The WebShop environment~\citep{webshop} offers a rich product dataset, requiring agents to analyze requirements, navigate multi-turn interactions, and demonstrate strong information retrieval, comparison, and decision-making skills in a complex web interface.
\end{itemize}

\subsection{Training and Evaluation Settings}
\textbf{Training.}\quad We leverage the Verl project as a foundation, implementing a fully asynchronous overhaul to develop a novel training framework, \model, tailored for agentic RL tasks. The framework was applied to train models including Qwen2.5-3B-Instruct, Qwen2.5-7B-Instruct, Qwen2.5-14B-Instruct, Qwen2.5-32B-Instruct, and GLM4-9B. The efficiency of the asynchronous design enabled extensive rollout training across the five selected multi-turn interaction tasks, facilitating large-scale RL with over 1000 steps in a multi-task mixed setting. This prolonged training ensured convergence of model performance across diverse tasks.

The interaction format between models and environments was standardized using the OpenAI Function Call Format. For the Qwen series, RL training commenced directly from the base models. In contrast, the GLM4-9B model required an initial cold-start phase with a limited set of supervised fine-tuning (SFT) data to adapt to the Function Call Format, followed by RL training, ultimately yielding significant performance improvements (see Table~\ref{tab:main}). Training was conducted on H800 GPUs, with a minimum configuration of 16 GPUs for the 14B model. Scalability was observed, as training efficiency increased with additional GPU resources.

The training process employed the Group Relative Policy Optimization (GRPO) algorithm as the baseline, enhanced with custom modifications (see Section~\ref{sec:multi-turn}). Rollouts were performed with a temperature of 0.8, sampling eight times per rollout to ensure diverse action exploration. To maintain consistency across multi-task environments, a binary reward function was designed, assigning a score based on the correctness of the entire trajectory. Trajectories exceeding the maximum interaction rounds or maximum response length incurred a penalty of -0.2. For computational efficiency, SGLang was adopted as the inference engine, paired with the Fully Sharded Data Parallel (FSDP) strategy to optimize RL training.

\textbf{Evaluation.}\quad For evaluation, a lightweight eval script was developed using the SGLang engine, seamlessly integrated with the asynchronous framework to enable rapid assessment of task performance. Evaluations were conducted with a temperature of 0.8, averaging results over four consecutive runs per task to ensure reliability. Additionally, a compatible API evaluation script was created to assess model APIs across tasks, supporting endpoints served by vllm or SGLang, with identical parameters (temperature 0.8, four-run average) to maintain consistency.

\subsection{Deployment Framework Details}
\label{sec:deploy-detail}

As shown in Figure~\ref{fig:overall-arch}, each worker in the new framework operates as a containerized execution unit, capable of managing concurrent task lifecycles under isolated runtime conditions. Workers are equipped with a detailed instrumentation layer for real-time observability, enabling telemetry at both session and task granularity. Internally, each worker integrates an abstract environment controller that mediates between task definitions and environment provisioning services. This controller is responsible for session instantiation, interaction handling, timeout enforcement, and environment cleanup. By abstracting the execution logic from physical deployment details, the worker layer can accommodate diverse backend configurations and support dynamic elasticity under shifting training loads.

The new controller adopts a non-blocking dispatch strategy that minimizes contention and ensures deadlock safety through a strict lock acquisition hierarchy. Timeout-driven fault detection and self-healing routines enable automatic de-registration and reintegration of unstable nodes. The controller also enforces strict lifecycle policies on session expiration, interaction timeout, and stale data cleanup through periodic maintenance loops.

\subsection{Results Analysis}

We provide a comprehensive evaluation of reinforcement learning (RL) performance across a diverse set of models and tasks. We report results for prominent API-based models and popular open-source base models. Additionally, we assess the RL-enhanced variants of our models at various scales, trained using the AgentRL framework. The evaluation extends to out-of-distribution (OOD) testing on an unseen benchmark, where the RL-trained model demonstrates performance gains over its base counterpart. Furthermore, we conduct an ablation study to investigate the impact of our proposed algorithmic techniques on model efficacy.

\textbf{Scaling Law} \quad The main results reflect a clear scaling law trend, with AgentRL-trained models showing consistent performance improvements as their size increases. Performance progressively escalates from the smallest model variants to the largest, indicating the framework's scalability and robustness. This progressive enhancement underscores the algorithm's adaptability to varying model sizes. The successful application to a model from a different architectural family further validates the framework's versatility, demonstrating its broad applicability beyond a single model series.

\textbf{Frontier Model Performance} \quad Comparative analysis highlights the superiority of our largest AgentRL-trained model over leading API-based models. While prominent proprietary LLMs achieve high scores, our RL-optimized model reaches a new state-of-the-art performance level, representing a substantial improvement over its base version before RL training. This suggests that AgentRL not only competes with but, in certain multi-turn and overall metrics, surpasses these advanced models, affirming its competitive edge.

\textbf{OOD Performance} \quad The OOD evaluation on the BFCL-v3 benchmark tests generalization on unseen tasks. The RL-trained model shows a clear improvement in overall performance compared to the base model, with a particularly significant leap in multi-turn task capability. This outperformance after extensive RL training underscores the method's ability to generalize beyond its training distribution, enhancing its potential for practical deployment in diverse scenarios.

\textbf{Ablation Study}\quad The ablation study further elucidates the efficacy of our methodological enhancements, detailed as follows:
\begin{itemize}[leftmargin=*,itemsep=0pt,parsep=0.2em,topsep=0.2em,partopsep=0.0em]
    \item \textbf{Cross-Policy Sampling}: This technique, designed to explore more states in open-ended environments, proves to be highly effective. Its inclusion boosts the average performance significantly. This result underscores the value of encouraging broader exploration, as the strategy successfully expands the model's capability boundaries by exposing it to more diverse and goal-relevant trajectories during training.
    
    \item \textbf{Task Advantage Normalization}: In contrast, this method stabilizes multi-task learning by mitigating negative interference and rate disparities across tasks. These findings support the selective integration of this technique, enhancing AgentRL's training stability and consistency.
\end{itemize}

\subsection{Case Studies}
\subsubsection{Case Study on the Efficacy of Cross-Sampling}
\begin{figure}[hbt]
    \centering
    \includegraphics[width=\linewidth]{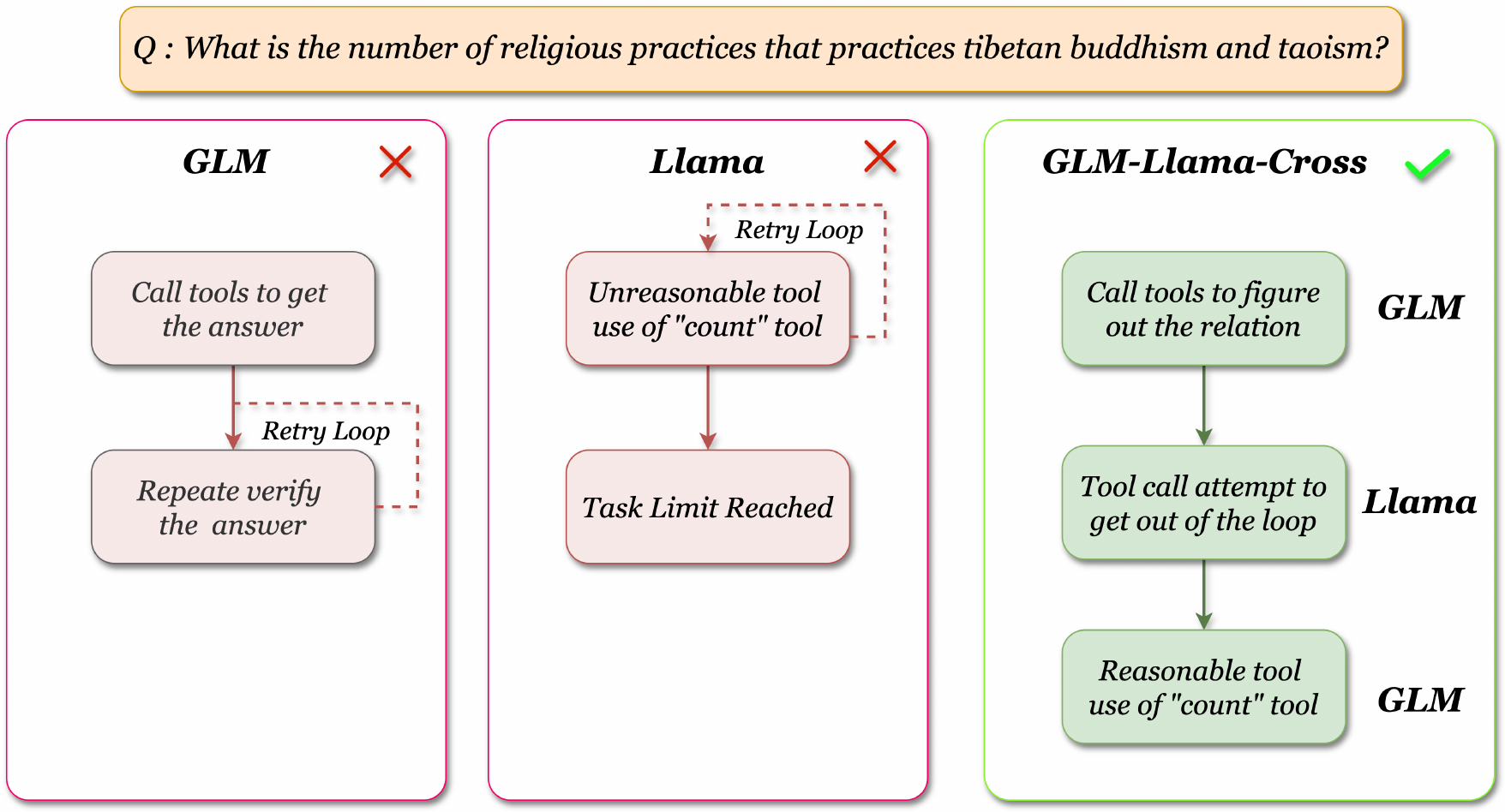}
    \caption{An example of GLM,Llama and GLM-Llama cross sampling in a KG task. This case study demonstrates the Cross-Policy Sampling strategy's success on a KG question-answering task, where GLM-4 fails in a conclusion loop and Llama falters with tool comprehension. It combines GLM-4's logic with Llama's tool interaction to achieve the correct answer.}
    \label{fig:cross-sample}
\end{figure}

To intuitively demonstrate the effectiveness of our proposed cross sampling strategy, we present a case study on a specific knowledge graph (KG) question-answering task. As shown in fig \ref{fig:cross-sample}, we analyze the execution trajectories of two models, \textbf{GLM-4-9B} and \textbf{Llama-8B}, on this task. The results show that when tasked individually, both models fail for different reasons. However, when applying our Cross-Policy  Sampling strategy, the agent successfully completes the task by finding the correct answer. 

The failures of the two individual models stem from distinct causes. \textbf{GLM-4} becomes trapped in a \textbf{premature conclusion loop}; it correctly deduces the final answer through logical inference but consequently bypasses the required protocol of using tools for verification. It repeatedly outputs its inferred conclusion in a non-standard format, leading to failure. In contrast, \textbf{Llama}'s failure is due to \textbf{flawed tool comprehension}; it persistently attempts to call tools with incorrect logic and parameters, indicating a fundamental misunderstanding of the tools' functionality and usage, which prevents any effective progress on the task.

The cross sampling strategy's success stems from a synergy that compensates for each model's weaknesses. It leverages GLM-4's strong logical planning to set a course, then breaks GLM-4's resulting non-interactive loop by switching to Llama's policy. Although Llama's own tool comprehension is flawed, its policy's critical function is to force an attempt at tool interaction. This switch to a "tool-centric" mode, guided by GLM-4's original logic, creates the opportunity for a valid tool call to emerge. This case study highlights  the superiority of Cross-Policy sampling by showing how it dynamically combines different problem-solving approaches to forge a successful path where single agents fail.

\subsubsection{Error Analysis}

We analyze the performance of the Qwen2.5-14B-Instruct model and the \model model across five environments (AlfWorld, DB, KG, OS, WebShop), focusing on the primary termination states: Completed and Task Limit Reached.

\begin{table*}[t]
    \centering
    \begin{tabular}{lcccc}
        \toprule
        \multirow{2}{*}{\textbf{Environment}} & \multicolumn{2}{c}{\textbf{Base Model}} & \multicolumn{2}{c}{\textbf{\model Model}} \\
        \cmidrule(lr){2-3} \cmidrule(lr){4-5}
        & \textbf{Completed} & \textbf{Task Limit Reached} & \textbf{Completed} & \textbf{Task Limit Reached} \\
        \midrule
        AlfWorld & 0.070 & 0.68 & 0.926 & 0.074 \\
        DB & 0.957 & 0.043 & 0.993 & 0.007 \\
        KG & 0.747 & 0.213 & 0.947 & 0.033 \\
        OS & 0.548 & 0.444 & 0.847 & 0.118 \\
        WebShop & 0.725 & 0.275 & 0.980 & 0.020 \\
        \bottomrule
    \end{tabular}
    \caption{Failure Modes Comparison. \textit{Note: "Completed" indicates the agent submitted an answer, not necessarily correctly, and these two statuses are not exhaustive; the sum of percentages may not reach 100\% due to other possible outcomes.}}
\end{table*}

The data highlights a substantial improvement with the \model method, where Completed rates increase significantly (e.g., from 0.070 to 0.926 in AlfWorld) and Task Limit Reached rates decrease (e.g., from 0.68 to 0.074 in AlfWorld). This suggests that RL training enhances the model's efficiency, reducing instances where tasks terminate due to time constraints and boosting successful completions across all environments.

\subsubsection{What Reinforcement Learning Teaches Models in ALFWorld}

We analyze a task from ALFWorld where the agent must place a saltshaker in a drawer. We compare the base model (Qwen2.5-14B-Instruct), which fails in four runs, with the RL-trained model (AgentRL-Qwen2.5-14B-Instruct), which succeeds in all four, to highlight RL's impact.

\textbf{Base Model Performance}\quad The base model struggles with:
\begin{itemize}[leftmargin=*,itemsep=0pt,parsep=0.2em,topsep=0.2em,partopsep=0.0em]
    \item \textbf{Improper Tool Usage}: Repeatedly attempts invalid actions (e.g., \texttt{look}) without using the \texttt{take\_action} tool, leading to errors.
    \item \textbf{Ineffective Strategy}: Fixates on cabinets (e.g., \texttt{cabinet 1}) without exploring likely locations like countertops, resulting in failure.
\end{itemize}

\textbf{RL-Trained Model Performance}\quad The RL-trained model excels by:
\begin{itemize}[leftmargin=*,itemsep=0pt,parsep=0.2em,topsep=0.2em,partopsep=0.0em]
    \item \textbf{Correct Tool Usage}: Consistently uses \texttt{take\_action} correctly, avoiding procedural errors.
    \item \textbf{Efficient Search}: Prioritizes countertops, quickly finding the saltshaker on \texttt{countertop 3}.
    \item \textbf{Action Sequencing}: Navigates to \texttt{drawer 1}, opens it, and places the saltshaker, completing the task.
\end{itemize}

From the above analysis we can see that reinforcement learning significantly enhances the model’s performance in ALFWorld by imparting tool proficiency for correct use of environment tools, strategic exploration to prioritize likely locations, and effective action planning for sequencing tasks, enabling efficient, goal-directed behavior that starkly contrasts with the base model’s repetitive failures.

\section{Prompt Examples}
\subsection{AlfWrold Task}
\begin{figure}[h]
\centering
\resizebox{\textwidth}{!}{
\begin{tcolorbox}[colback=gray!5!white, colframe=blue!75!black, 
title=System Prompt for AlfWorld, boxrule=0.3mm, width=\textwidth, arc=3mm, auto outer arc=true, fontupper=\small]

Interact with a household to solve a task. Imagine you are an intelligent agent in a household environment and your target is to perform actions to complete) the task goal. 

At the beginning of your interactions, you will be given the detailed description of the current environment and your goal to accomplish. A tool will be provided for you to use to submit the action you want to take. This tool is the only tool you should and must take in order to operate any action in the environment. The way you perform action is to place the action chosen by you in the arguments field of your tool call. 

For each of your turn, you will be given a list of actions which you can choose one to perform in this turn. The action you would like to take should be offered in this format: \"the name of your next action\", and you should fill it in the argument field of your tool call. Note that you should always call a tool to operate an action from the given choices. After your each turn, the environment will give you immediate feedback based on which you plan your next few steps. if the environment output \"Nothing happened\", that means the previous action is invalid and you should try more options.

\textbf{Reminder:}
\begin{itemize}[leftmargin=*,itemsep=0pt,parsep=0.2em,topsep=0.2em,partopsep=0.0em]
    \item the action must be chosen from the given available actions. Any actions except provided available actions will be regarded as illegal.
    \item Always call the tool to hand in your next action and think when necessary.
\end{itemize}
\end{tcolorbox}
}
\label{prompt:alfworld}
\end{figure}

\subsection{Knowledge Graph (KG) Task}

\begin{figure}[h]
\centering
\resizebox{\textwidth}{!}{
\begin{tcolorbox}[colback=gray!5!white, colframe=blue!75!black, 
title=System Prompt for Knowledge Graph, boxrule=0.3mm, width=\textwidth, arc=3mm, auto outer arc=true]
\textbf{Instructions:}
You are an intelligent agent tasked with answering questions based on the knowledge 
stored in a \textbf{knowledge base (KB)}. Utilize the provided tools to probe the KB 
and retrieve relevant information to address user queries effectively.

Navigate the KB to identify \textbf{relationships}, \textbf{attributes}, and \textbf{intersections}. 
where applicable, ensuring the most pertinent information is used to formulate answers.

\textbf{Remember:}
\begin{itemize}[leftmargin=*,itemsep=0pt,parsep=0.2em,topsep=0.2em,partopsep=0.0em]
    \item A variable can be an entity or a set of entities resulting from previous queries.
    \item Ensure the tool selected aligns with the question's demands, following a 
          logical order (e.g., fetch relations before finding neighbors).
    \item After generating a variable, assess whether it constitutes the \textbf{final answer}. 
          Variables are assigned IDs starting from 0 (e.g., \#0, \#1, etc.).
    \item Upon identifying the \textbf{final answer}, respond with 'Final Answer: \#id', 
          where \#id is the variable's ID (e.g., 'Final Answer: \#3'). Do not invoke tools 
          after determining the final answer!
    \item Execute one action at a time, with a maximum of 15 actions to find the answer.
    \item Use the supplied tools unless the \textbf{final answer} is identified.
\end{itemize}

Your thoughtful application of these tools and careful consideration of interactions 
will guide you to correct answers. Note that the task must be completed within 15 rounds— 
plan your attempts accordingly!

\end{tcolorbox}
}
\label{prompt:kg}
\end{figure}

\clearpage
\subsection{DB Task}
\begin{figure}[h]
\centering
\resizebox{\textwidth}{!}{
\begin{tcolorbox}[colback=gray!5!white, colframe=blue!75!black, 
title=System Prompt for DataBase, boxrule=0.3mm, width=\textwidth, arc=3mm, auto outer arc=true, fontupper=\small]

I will ask you a question, then you should help me operate a \textbf{MySQL database} with SQL to answer the question.You have to explain the problem and your solution to me and write down your thoughts.After thinking and explaining thoroughly, every round you can choose to \textbf{operate or to answer} with the two specific tools provided.

If you should execute a SQL query, use the `execute\_sql` function, Your SQL should be in one line. Every time you can only execute one SQL statement. I will only execute the statement in the first SQL code block. Every time you write a SQL, I will execute it for you and give you the output. If you are done operating, and you want to commit your final answer, then use the \`commit\_final\_answer` function.

DO NOT use this tool unless you are sure about your answer. I expect an accurate and correct answer.Your answer should be accurate. Your answer must be exactly the same as the correct answer.If the question is about modifying the database, then after done operation, your answer field can be anything.If your response cannot match any pattern I mentioned earlier, you will be judged as FAIL immediately.You should always use the tools provided to submit your answer. Be careful not to write it in the content field.Your input will be raw MySQL response, you have to deal with it by yourself.

\end{tcolorbox}
}
\label{prompt:db}
\end{figure}

\subsection{OS Task}
\begin{figure}[h]
\centering
\resizebox{\textwidth}{!}{
\begin{tcolorbox}[colback=gray!5!white, colframe=blue!75!black, 
title=System Prompt for Operating System, boxrule=0.3mm, width=\textwidth, arc=3mm, auto outer arc=true, fontupper=\small]

You are an assistant that will act like a person. I will play the role of a \textbf{Linux (Ubuntu) operating system}. Your goal is to implement the operations required by me or answer the questions proposed by me.

For each of your turns, you should first think about what you should do, and then call exactly one of the provided tools according to the situation.If you think the output is too long, I will truncate it. The truncated output is not complete. You have to deal with the truncating problem by yourself.

\textbf{Attention}, your bash code should not contain any input operation. Once again, you should use one tool in each turn, and should not respond without function calling.

Note that if you think the task has been finished, or there is some message missing to completely complete the task, you should respond with calling the function \"finish\_action\", as no additional information will be provided.

Also, note that if you have gotten the answer to the question, you should call the \"answer\_action\" tool instead of simply writing your answer in your response.

Your answers should be exact and precise (for example, a single number), do not answer with full sentences or phrases.Always use a tool provided instead of simply responding with content.

\end{tcolorbox}
}
\label{prompt:os}
\end{figure}

\subsection{Webshop Task}

\begin{figure}[h]
\centering
\resizebox{\textwidth}{!}{
\begin{tcolorbox}[colback=gray!5!white, colframe=blue!75!black, 
title=System Prompt for Web Shopping, boxrule=0.3mm, width=\textwidth, arc=3mm, auto outer arc=true, fontupper=\small]
You are web shopping. I will provide \textbf{instructions} about what to do, and you must follow them strictly.

Every round, you will receive an observation and a list of available actions. You must respond by calling a tool based on the current state and instructions.

\begin{itemize}[leftmargin=*,itemsep=0pt,parsep=0.2em,topsep=0.2em,partopsep=0.0em]
    \item You can use the \textbf{search tool} if it is available.
    \item You can click one of the buttons in \textbf{clickables}.
    \item If an action is not valid, perform nothing.
\end{itemize}

Keywords for the \textbf{search tool} are your choice, but the value for a click must be from the list of available actions. Remember to design search keywords carefully.

First, think about what to do, then call a tool accordingly. You should always use a tool, even if you have questions to confirm, and you can use any available tool without user permission.

\end{tcolorbox}
}
\label{prompt:webshop}
\end{figure}

\section{Discussions}
\label{sec:discussion}

\subsection{Limitations}
\label{subsec:limitations}

While our framework establishes a new state-of-the-art in agentic RL, we identify two primary areas for future research that build upon our solid foundation. First, our novel cross-policy sampling strategy is a key driver of enhanced exploration. By its very design of integrating diverse policies, it can introduce minor distributional shifts. These shifts can manifest as mild, transient instabilities in training dynamics, a manageable trade-off for achieving broader state-space coverage. Future work could explore principled refinements, such as adaptive policy weighting, to further optimize this powerful mechanism. Second, as a foundational work, this paper focuses on rigorously validating our framework across a comprehensive suite of controlled environments. Having established the system's robustness and scalability, the natural next step is its application to more complex and dynamic real-world scenarios. We believe our framework provides the ideal testbed for tackling this exciting challenge.

\subsection{Future Works}

Looking ahead, we plan to extend \model to a broader range of environments and scale it to larger models. Future research will also explore more sophisticated variants of cross-policy sampling and develop improved methods for multi-task optimization. We believe these are crucial steps toward creating more general and capable LLM agents.

\section{Use of LLMs}
During the preparation of this manuscript, we used large language models (LLMs) to assist with 
language polishing and grammar improvement. All research ideas, methods, experiments, and 
analyses were conceived, designed, and validated by the authors.

\end{document}